\documentclass[11pt,a4paper]{article}
\usepackage[margin=1in]{geometry}
\usepackage[T1]{fontenc}
\usepackage[utf8]{inputenc}
\usepackage[nodisplayskipstretch]{setspace}
\usepackage{amsmath,amsfonts,amssymb,amsthm,mathtools}
\usepackage{subfigure,dsfont,booktabs,enumerate}
\usepackage[table,dvipsnames]{xcolor}
\usepackage[most]{tcolorbox}
\usepackage{multicol,multirow,graphicx,wrapfig,float}
\usepackage[authoryear,round]{natbib}
\usepackage[ruled]{algorithm2e}
\SetKwInput{KwInput}{Input}
\SetKwInput{KwOutput}{Output}
\usepackage{authblk}
\usepackage[hidelinks]{hyperref}
\usepackage{bookmark,cleveref}
\newcommand{\argmax}{\operatorname*{arg\,max}}
\newcommand{\argmin}{\operatorname*{arg\,min}}

\newtheorem{theorem}{Theorem}
\newtheorem{proposition}{Proposition}

\newtheorem{corollary}{Corollary}
\newtheorem{definition}{Definition}
\newtheorem{assumption}{Assumption}

\crefname{assumption}{assumption}{assumptions}
\Crefname{assumption}{Assumption}{Assumptions}
\title{Black-Box Test-Time Personalization: Optimality and Collaboration}
\author[1]{Yuheng Ma}
\author[2]{Feiyu Jiang}
\author[3]{Zifeng Zhao}
\author[4]{Yi Yu}
\affil[1]{KLATASDS-MOE, School of Statistics, East China Normal University, yhma@sfs.ecnu.edu.cn.}
\affil[2]{School of Management, Fudan University, jiangfy@fudan.edu.cn.}
\affil[3]{Mendoza College of Business, University of Notre Dame, zifeng.zhao@nd.edu.}
\affil[4]{Department of Statistics, University of Warwick, yi.yu.2@warwick.ac.uk.}

\date{}

\begin{document}
\maketitle
\begin{abstract}

Personalizing a frozen black-box language model is difficult when no reliable reward model or verifier is available. We study online best-of-$K$ response selection: at each interaction, the model generates a size-$K$ candidate set, among which the system chooses one response to present. By formulating this task as a nonparametric stochastic  contextual bandit problem, we propose Black-Box Test-Time Nearest Neighbors, \texttt{(BTN)}\textsuperscript{2}, an adaptive nearest-neighbor upper confidence bound policy that smartly learns user preferences from sequential feedback. Matching regret upper and lower bounds are established up to logarithmic factors, revealing a phase transition in the candidate-set size~$K$.  To accelerate learning across multiple users, we extend this framework to a collaborative bandit setting and characterize user similarity through the distributions of preferred choices, which allows effective information sharing among users with the same preference ordering despite potentially different reward models.
Based on this similarity measure, we propose a two-stage procedure that detects latent user clusters and pools feedback for joint decision-making.    Theoretical guarantees on cluster recovery and collaborative regret are  established.  Extensive numerical experiments are also provided, including simulations examining the predicted rates and collaboration effects, and a question-answering benchmark with constructed user preferences showing the superiority of our methods over other state-of-the-art baselines.
\end{abstract}

\section{Introduction}

The performance of a large language model (LLM) depends not only on its learned parameters, but also on the procedure used to generate its responses.
Test-time computation offers a practical way to improve LLM responses at inference time \citep{snell2025scaling}. Unlike pretraining or post-training, it leaves the parameters of the base model fixed and instead spends computation on sampling alternative responses \citep{cobbe2021training}, searching over reasoning trajectories \citep{zhou2024language}, or iteratively revising the output \citep{madaan2023selfrefine}. 
This framework is well suited when we have access to a powerful base model but cannot update its parameters because of computational constraints \citep[e.g., a large open-weight model,][]{deepseekai2026deepseek} or black-box restrictions \citep[e.g., a proprietary model,][]{openai2023gpt}, yet still require improved performance or personalized responses.

Best-of-$K$ sampling is a widely used test-time strategy for improving  outputs of black-box generative models, especially LLMs, on complex tasks by increasing the chance of obtaining a high-quality response \citep{stiennon2020learning,cobbe2021training,snell2025scaling}. For each query, it samples $K$ candidate responses from the model and returns the one with the highest score. A standard best-of-$K$ selection ranks candidates using either a programmatic verifier when correctness can be checked automatically, as in some math or coding tasks \citep{li2022competition}, or a proxy reward model for open-ended tasks without a reliable verifier \citep{nakano2021webgpt}. Without such a verifier or proxy reward model, selecting the best response from a candidate set becomes difficult \citep{qu2025tpop}.

\begin{figure}[!t]
    \centering
    \includegraphics[width=\textwidth]{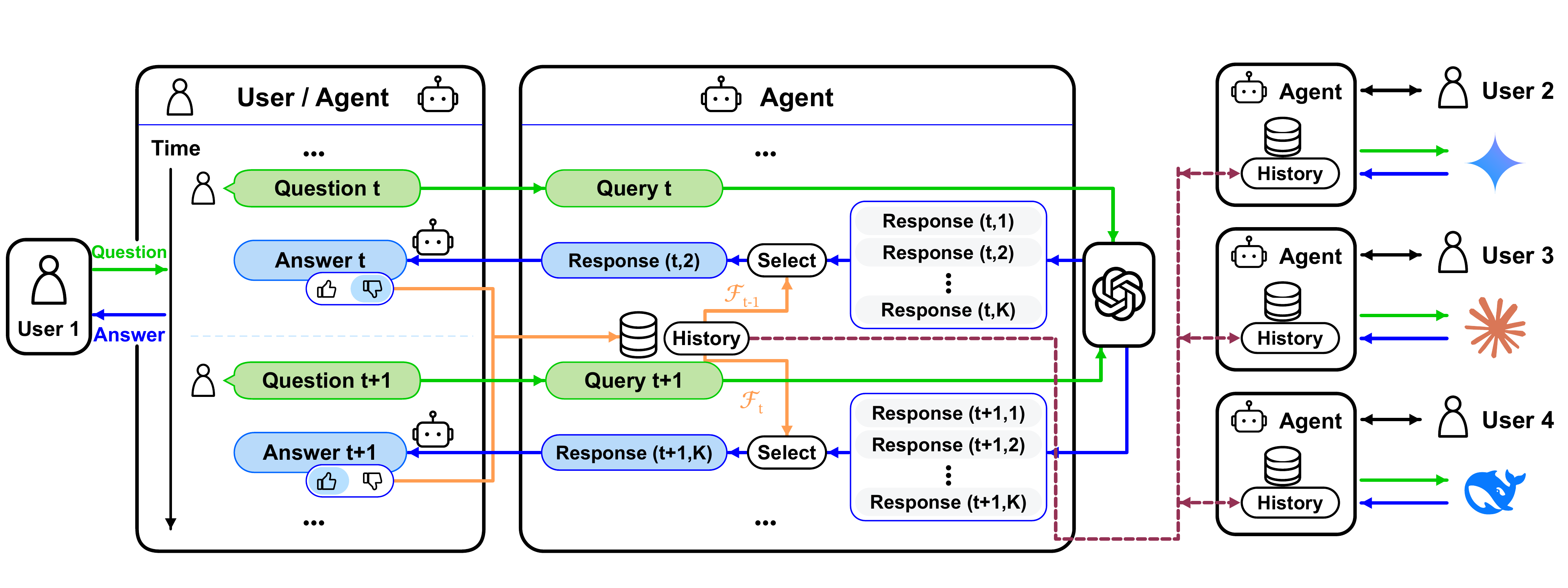}
    \caption{Illustration of the online interaction process. Left and center (single-user personalization): at round $t$, the user submits a question and the agent queries a black-box LLM to obtain $K$ candidate responses; using the prior history $\mathcal{F}_{t-1}$ (orange arrows), the agent selects one response to return and records the user's feedback to update the history to $\mathcal{F}_t$ for subsequent rounds. Right (multi-user collaboration): multiple users interact with their personal agents in parallel, and agents exchange historical feedback (dashed lines) across users with compatible preferences to accelerate learning. }
    \label{fig:interaction-process}
\end{figure}

We study test-time personalization in an online setting, where responses from a black-box model are adapted to individual users through sequential feedback. %
The interaction process is illustrated in Figure~\ref{fig:interaction-process}.
Consider an AI agent system deployed on personal devices that obtains responses from LLMs using user-provided API keys.
The system combines these responses with its own capabilities to deliver services to users.
At each interaction, the LLM produces $K$ candidate responses whose query--response representations form a candidate set.
The agent system returns only one response  to the user and observes the resulting feedback. %
This setting presents two challenges. First, each user has an unknown and potentially complex reward function reflecting their own preferences, which the agent must learn efficiently.
Second, to achieve accelerated learning, a natural strategy is to further leverage information across different users, which then requires identifying users whose preferences are sufficiently compatible to enable effective knowledge transfer.

We first decompose the cumulative utility loss relative to the globally optimal response into generation and selection components.
The generation gap is the {utility loss} between the global optimum and the best generated candidate, while the selection gap is the loss between that candidate and the response returned.
With the base model fixed, the agent can control only the selection gap, which it seeks to reduce by learning from feedback.
We therefore formulate the minimization of the selection gap as a stochastic contextual bandit problem \citep{abbasiyadkori2011improved,bubeck2011xarmed}.
Given multiple users, we view feedback aggregation as a clustered bandit problem \citep{gentile2014online,miao2022context}.
We summarize our contribution as follows.
\begin{enumerate}[(a)]
    \item Section~\ref{sec-single-user} formulates the single-user best-of-$K$ selection problem as a bandit problem and develops \texttt{(BTN)}\textsuperscript{2}, a nearest-neighbor (NN) upper confidence bound (UCB) policy.
    Theoretically, we establish matching upper and lower bounds, characterize a phase transition in the number of candidates, and extend the analysis to intrinsic low-dimensional structure.
    \item Section~\ref{sec:collaboration} formulates the multiple-user best-of-$K$ selection problem as a collaborative bandit clustering problem and defines a user similarity score based on the distributions of users' oracle choices. 
    This score captures preference similarity despite potential differences in reward models, providing a principled basis for feedback sharing. 
    We develop a two-stage method comprising cluster detection and collaborative learning, supported by theoretical guarantees for cluster recovery and collaborative regret that characterize the benefits of collaboration.
    \item Extensive numerical experiments validating the theoretical findings are collected in Section~\ref{sec:experiments}, accompanied with a real-world experiment on a question-answering benchmark demonstrating the superiority of the proposed methods over the state-of-the-art baselines.

    \item We extend the framework to accommodate more realistic interactions, allowing for varying candidate-set sizes  across queries, dependent responses generated within each query, and asynchronous user arrivals.
    These extensions are presented in Appendix~\ref*{app:extensions}.
\end{enumerate}

\noindent \textbf{Notation.} Let $a_n \lesssim b_n$ and $a_n \gtrsim b_n$ denote that there exist positive constants $n_1 \in \mathbb{N}$, $c, c' > 0$ such that $a_n \leq c b_n$ and $a_n \geq c' b_n$, for all $n \geq n_1$, respectively. We denote $a_n\asymp b_n$ if $a_n\lesssim b_n$ and $b_n\lesssim a_n$.
We write $a\lesssim_{\log} b$ or $a \asymp_{\log} b$ when $a \lesssim b$ or $a \asymp b$ up to polylogarithmic factors.
Let $a\vee b = \max (a,b)$ and $a\wedge b = \min (a,b)$.
Let $f_1 \circ f_2$ represent the composition of functions $f_1$ and $f_2$.
A closed Euclidean ball centered at $x$ with radius $r\in(0,+\infty)$ is denoted by $B(x,r):=\{z:\|z-x\|_2\leq r\}$.
For any positive integer $K$, denote $[K]=\{1,2,\ldots,K\}$.

\section{Best-of-$K$ Selection} \label{sec-single-user}

In this section, we formulate the single-user best-of-$K$ selection problem in \Cref{sec:nonparametric-settings}, introduce \texttt{(BTN)}\textsuperscript{2} with its ambient-dimension regret guarantees in \Cref{sec:adaptive-knn}, and extend the analysis to accommodate low-dimensional structure in \Cref{sec:intrinsic-dimension}.  
Further discussions on the role of the candidate-set size through the lens of generation and selection gaps are provided in \Cref{sec:global-regret}.

\subsection{Settings}
\label{sec:nonparametric-settings}

We first introduce the settings for the nonparametric stochastic bandit problem. 
In what follows, the bandit formulation does not require distinguishing between the user and its agent. For clarity, we refer to both collectively as the user, who observes candidates, selects actions, and updates the policy.
Let $\mathcal{X}\subseteq\mathbb{R}^d$ be a nonempty compact set with Euclidean diameter at most one.
For $t \in \mathbb{N}$, let~$\mathcal{F}_t$ be the $\sigma$-field generated by the candidate sets, selected actions, rewards, and policy randomization revealed up to time $t$, with the convention that $\mathcal{F}_0 = \sigma(\varnothing)$.
In the single-user model, at each round $t$, 
the candidate set $\mathcal{D}_t=\{X_{t,j}:j\in[K]\}$ consists of $K \geq 2$ independent and identically distributed (i.i.d.) draws from $\mathrm{P}_X$, independent of $\mathcal{F}_{t-1}$. Both independence and a fixed $K$ are assumed for convenience. 
{
Our results can be readily extended to settings in which $X_{t,j}$'s are identically distributed but not necessarily independent, or in which $K$ varies across time steps.
See Appendices~\ref*{app:varying-k} and~\ref*{app:correlated-candidates} for these extensions.}
Choosing $X_{t}\in \mathcal{D}_t$ yields a reward $Y_t\in[0,1]$ satisfying 
\begin{align}
\mathbb{E}\big[Y_t\,|\,\mathcal{F}_{t-1},\mathcal{D}_t,X_t\big]=f(X_t), 
    \label{eq:history-conditional-reward}
\end{align}
where $f:\mathcal{X} \to [0,1]$ is the reward function.
{
Throughout, we assume that $f$ is Lipschitz continuous, i.e., there exists a constant $C_L>0$ such that $|f(x)-f(x')| \leq C_L\|x-x'\|_2$ for all $x,x'\in\mathcal{X}$.}
Let $\mathcal{Z}_t:=\{(X_s,Y_s)\}_{s = 1}^t$ denote the sample history up to time $t$, with $\mathcal{Z}_0=\varnothing$.
The policy at time $t$ is nonanticipating and only based on $\mathcal{F}_{t-1}$, writing as $X_t=\pi_t(\mathcal{D}_t)$, omitting the notational dependence on $\mathcal{F}_{t-1}$ and the policy's internal randomization.
To the best of our knowledge, this nonparametric stochastic bandit problem has not been studied in the literature. Existing work either assumes a linear reward function  $f$ \citep{abbasiyadkori2011improved} or allows actions to be chosen freely from $\mathcal{X}$ rather than restricting them to the candidate set $\mathcal{D}_t$ \citep{bubeck2011xarmed}.
As $K\to\infty$, the candidate set becomes increasingly dense in $\mathcal{X}$, providing an accurate approximation to the full action space considered in continuum-armed bandits \citep{auer2007improved,bubeck2011xarmed}.
This setting models online best-of-$K$ response selection for a black-box LLM, as illustrated in Figure~\ref{fig:interaction-process}.
As discussed in the introduction, relative to the best response over the entire response space, the reward gap of the selected response admits the exact decomposition
\begin{align}
\label{equ:generation-selection-gap}
\sup_{x\in\mathcal{X}} f(x)-f(X_t)
=& \underbrace{\sup_{x\in\mathcal{X}} f(x)-\max_{x\in\mathcal{D}_t}f(x)}_{\text{generation gap}}
+ \underbrace{\vphantom{\sup_{x\in\mathcal{X}}}\max_{x\in\mathcal{D}_t}f(x)-f(X_t)}_{\text{selection gap}}.
\end{align}
The generation gap measures the difference between the globally best response and the best generated one. It is governed by the capability of the base model, the prompt, and the sampling and decoding procedure. Our analysis takes these as fixed and focuses on the selection gap: the difference between the best response in the realized candidate set and the response actually selected.  
We therefore evaluate a policy against an oracle that observes the same candidate set $\mathcal{D}_t$ and knows the reward function $f$.

Denote $\pi^{*}$ as the oracle policy, i.e.,~$\pi^*(\mathcal{D})\in\argmax_{X\in\mathcal{D}}f(X)$. The main objective is to design a policy $\pi = \left\{\pi_{t}\right\}_{t=1}^T$ minimizing the candidate-set regret defined as
\begin{align}
    R(\pi) = \sum_{t=1}^{T} \mathbb{E}_{\mathcal{D} \sim \mathrm{P}_X^{\otimes K}}\left[ f(\pi^{*}(\mathcal{D}))-f(\pi_{t}(\mathcal{D})) \mid \mathcal{F}_{t - 1} \right]. \label{equ:def-regret-single}
\end{align}
Let $X^*_{t} = \pi^{*}(\mathcal{D}_t)$. We remark that in \eqref{equ:def-regret-single}, each summand is the instantaneous candidate-set regret of policy $\pi_{t}$, where the expectation is taken with respect to the decision set $\mathcal{D}$, which is independent of $\mathcal{F}_{t - 1}$.

\subsection{Nearest-Neighbor Upper Confidence Bound}
\label{sec:adaptive-knn}

\subsubsection{Methodology}

We now introduce Black-Box Test-Time Nearest Neighbors, \texttt{(BTN)}\textsuperscript{2}, a nearest-neighbor algorithm for best-of-$K$ selection that gradually approaches the oracle policy $\pi^*$ above.
At each round $t$, \texttt{(BTN)}\textsuperscript{2} constructs a $k$-NN-based UCB for each candidate $x\in\mathcal{D}_t$ using $\mathcal{Z}_{t-1}$, with the number of neighbors $k$ selected adaptively for each candidate.
It then selects the candidate $X_t\in\mathcal{D}_t$ with the largest UCB.
For a fixed time $t$, consider the history $\mathcal{Z}_t$ and a query point $x\in\mathcal{X}$.
Let $(\sigma_s(x; \mathcal{Z}_t))_{s = 1}^t$ list the indices in $[t]$ in nondecreasing order of $\{\|X_s - x\|_2\}_{s = 1}^t$.
For $k\in[t]$, define the $k$-NN set $\mathcal{I}_k(x;\mathcal{Z}_t):= \{\sigma_j(x;\mathcal{Z}_t):j\in[k]\}$ and the $k$-distance $r_k(x;\mathcal{Z}_t):= \|X_{\sigma_k(x, \mathcal{Z}_t)}-x\|_2$.
Based on $\mathcal{Z}_t$, with a pre-specified $k$, a $k$-NN estimator of the reward function is
\begin{align} \label{eq-knn-ind}
\widehat f_k(x;\mathcal{Z}_t) 
:=
\frac{1}{k} \sum_{s\in\mathcal{I}_k (x; \mathcal{Z}_t) }Y_s.
\end{align}
{The pointwise error $|\widehat f_k-f|$ of \eqref{eq-knn-ind} arises from two sources \citep{kpotufe2011knn}. One is the approximation error, which arises from local variation of $f$ and, under the Lipschitz condition, is bounded by a constant multiple of $r_k(x;\mathcal{Z}_t)$. The other is the estimation error, which arises from averaging $k$ noisy rewards and is of order $k^{-1/2}$.
This motivates defining the confidence radius as a surrogate for the pointwise error:
\begin{align}
\label{eq:def-ucb-radius}
U_k(x,\mathcal{Z}) := C r_k(x;\mathcal{Z})+\sqrt{\frac{\ell}{k}},
\end{align}
where $C>0$ and $\ell>0$ are tuning parameters.
Following \citet{reeve2018k}, we select the   tuning parameter $k$ using an adaptive procedure. 
Specifically, for $t\geq2$, choose
\begin{align}
\label{eq:def-adaptive-knn-k}
\widehat k_t(x) 
\in
\argmin_{k\in[t-1]} U_k(x, \mathcal{Z}_{t-1})
\end{align}
and add the minimum confidence radius to the corresponding $k$-NN estimate to obtain
\begin{align}
\label{eq:def-ucb}
\operatorname{UCB}_t(x;\mathcal{Z}_{t-1})
:=
\widehat f_{\widehat k_t(x)}(x;\mathcal{Z}_{t-1})
+
U_{\widehat k_t(x)}(x,\mathcal{Z}_{t-1}),
\end{align}
with $\operatorname{UCB}_1(x;\mathcal{Z}_0) = \infty$.
}
Finally, at round $t$, the policy selects
\begin{align}
\label{equ:def-policy-adaptivenn}
X_t
\in
\argmax_{x\in\mathcal{D}_t}
\operatorname{UCB}_t(x;\mathcal{Z}_{t-1}).
\end{align}
After observing the reward $Y_t$, we append $(X_t, Y_t)$ to $\mathcal{Z}_{t-1}$ to form $\mathcal{Z}_t$.
We illustrate how the confidence bounds of \texttt{(BTN)}\textsuperscript{2} evolve with time in Appendix~\ref*{app:adaptivenn-details}.

\subsubsection{Theoretical Guarantee} \label{sec-ambient-dim-regret}

We impose the following two regularity assumptions.

\begin{assumption}
\label{def:lower-mass}
There exists a constant $\underline{c}>0$ such that $\mathrm{P}_X(B(x,r)) \geq \underline{c}r^d$ for all $x\in\mathcal{X}$ and $r\in(0,1]$.
\end{assumption}

\begin{assumption}
\label{def:uc-beta} 
Let $X\sim \mathrm{P}_X$.
Assume that $f(X)$ satisfies Anti-Concentration-$\beta$ (AC-$\beta$), i.e., there exist an exponent $\beta\in(0,1]$ and a constant $C_{\operatorname{AC}}>0$ such that $\sup_{u\in\mathbb{R}}\mathbb{P}\!\left(u<f(X)\le u+b\right)\le C_{\operatorname{AC}}b^\beta$ for all $b\in(0,1]$. %
\end{assumption}

{
\Cref{def:lower-mass} is  standard in nonparametric regression literature \citep{gyorfi2002distribution}. 
\Cref{def:uc-beta} controls the concentration of $f(X)$ by limiting the probability mass in any short utility interval.
It is analogous to a margin condition in classification \citep{audibert2007fast}, but controls concentration uniformly over all utility levels.
The case $\beta=1$ corresponds to a bounded density, whereas smaller $\beta$ allows stronger concentration.
Let $\mathcal{F}_{\operatorname{AC}}(d,\beta)$ denote the class of distributions in dimension $d$ satisfying \eqref{eq:history-conditional-reward}, with $f$ being Lipschitz continuous, $\mathrm{P}_X$ satisfying \Cref{def:lower-mass}, and $f(X)$ satisfying \Cref{def:uc-beta} with exponent $\beta$.
}

We first provide an upper bound on the cumulative regret \eqref{equ:def-regret-single} achieved by \texttt{(BTN)}\textsuperscript{2} and then establish a matching minimax lower bound, showing that the dependence on $K$ and $T$ is optimal up to logarithmic factors.

\begin{theorem}[Regret upper bound]
\label{thm:adaptive-upper}
Consider any problem instance $\mathrm{P}\in\mathcal{F}_{\operatorname{AC}}(d,\beta)$ with candidate-set size $K$ and time horizon $T$.
Let $\pi_{\mathtt{(BTN)}^2}$ be the \texttt{(BTN)}\textsuperscript{2} policy specified in \eqref{equ:def-policy-adaptivenn}, with $C$ and $\ell=c_0\log(KT)$, where $C$ and $c_0$ are sufficiently large constants depending only on the distribution.
It then holds that
\begin{align}
\label{eq:adaptive-rate}
\mathbb{E}[R(\pi_{\mathtt{(BTN)}^2})]
\lesssim_{\log}
(KT)^{\frac{d+1-\beta}{d+2}}
\wedge
T^{\frac{d+1-\beta}{d+2-\beta}}. 
\end{align}
\end{theorem}

\begin{theorem}[Lower bound]
\label{thm:lower-bound}
For every $d\geq1$ and $\beta\in(0,1]$, for all $K\geq2$, we have
\begin{align*}
\inf_\pi \sup_{\mathrm{P}\in\mathcal{F}_{\operatorname{AC}}(d,\beta)} \mathbb{E}_{\mathrm{P}}[R(\pi)]
\gtrsim
(KT)^{\frac{d+1-\beta}{d+2}}
\wedge
T^{\frac{d+1-\beta}{d+2-\beta}}.
\end{align*}
\end{theorem}

{
The minimax rate has two regimes, corresponding to $(KT)^{\frac{d+1-\beta}{d+2}}$ and $T^{\frac{d+1-\beta}{d+2-\beta}}$, with a phase transition at $K\asymp T^{\frac{\beta}{d+2-\beta}}$. 
When $K\lesssim T^{\frac{\beta}{d+2-\beta}}$, the regret grows as $(KT)^{\frac{d+1-\beta}{d+2}}$, where $K$ and $T$ together contribute as effective sample size.
In this regime, increasing $K$ produces a denser candidate set and smaller utility gaps among the leading candidates, making it harder to identify the oracle candidate and thereby leading to a larger regret. 
When $K\gtrsim T^{\frac{\beta}{d+2-\beta}}$, the second regime dominates, and the regret saturates at $T^{\frac{d+1-\beta}{d+2-\beta}}$: intuitively, the candidate set is so dense that distinguishing closely competing candidates requires finer accuracy than is attainable when estimating a Lipschitz $f$ from observed samples, so further increasing $K$ does not change the rate.
In general, this rate concerns only the selection gap and is nondecreasing in $K$. However, a larger $K$ can reduce the generation gap and improve the overall performance.
See Section~\ref{sec:global-regret} and Appendix~\ref*{app:derivation-generation-gap} for further discussions.}

{
Observing the rate, a larger $\beta$ decreases both regret exponents, while moving the phase transition to a larger value of $K$. 
This parallels the role of margin conditions in classification, where controlling the probability mass near the decision boundary leads to faster excess-risk rates \citep{audibert2007fast}.
Here, stronger anti-concentration provides tighter control over the probability of encountering suboptimal candidates whose rewards are close to that of the best candidate, thereby improving the rates for the cumulative selection gap.}
{We next relate our regret rates to the literature.
Nonparametric bandits with randomly sampled candidate sets have rarely been studied, making it difficult to compare the first term with existing results. We therefore focus on the second term.
When $\beta=0$, this term reduces to $T^{(d+1)/(d+2)}$, which is the canonical minimax order for $d$-dimensional Lipschitz bandits \citep{kleinberg2004nearly,kleinberg2008multi,bubeck2011lipschitz}.
More generally, our rate is compatible with the $X$-armed  bandit \citep{bubeck2011xarmed} form $(D+1)/(D+2)$ with effective near-optimality dimension $D=d-\beta$, reflecting how the anti-concentration assumption reduces the mass geometry of near-optimal regions. This correspondence is also corroborated by the one-dimensional continuum-armed setting: under local Lipschitz smoothness and a near-optimal-set mass condition with exponent~$\beta$, \citet{auer2007improved} obtain $\widetilde{O}\bigl(T^{(2-\beta)/(3-\beta)}\bigr)$, which equals our saturated rate for $d=1$.} 

\subsection{Utilizing Low-Dimensional Structure}
\label{sec:intrinsic-dimension}

Language-model responses are often mapped to extremely high-dimensional hidden-state embeddings for downstream comparison or selection (e.g.~$7168$ dimensions in \citealt{deepseekai2026deepseek} and $16384$ in \citealt{grattafiori2024llama}), leading to near-linear regret as demonstrated in  \Cref{thm:lower-bound}.
Recent research, fortunately, provided empirical evidence of low-dimensional geometric structure in language-model representations \citep{cai2021isotropy,cheng2023bridging}.
This motivates us to exploit a low, intrinsic dimensional structure, formalized in \Cref{def:intrinsic-dimension}.%

\begin{assumption}
\label{def:intrinsic-dimension}
There exist an intrinsic-dimension upper bound $d_*\in(0,d] \cap \mathbb{Z}$ and a constant $\underline{c}_*>0$ such that $\mathrm{P}_X(B(x,r)) \geq \underline{c}_* r^{d_*}$ for all $x\in\mathcal{X}$ and $r\in(0,1]$.
\end{assumption}

Compared to \Cref{def:lower-mass}, \Cref{def:intrinsic-dimension} enforces an lower bound on probability mass in a low-dimensional space.  It ensures that $\mathcal{X}$ can be covered by a collection of  dimension-$d^{*}$ balls of radius $r$.
This is a mild condition which can be implied by commonly used conditions, 
including compact manifold assumptions \citep{bickel2007local,cheng2013local} and the global homogeneity condition \citep{kpotufe2011knn}.

Let $\mathcal{F}_{\operatorname{AC}}(d,d_*,\beta)$ denote the class of distributions in ambient dimension $d$ satisfying \eqref{eq:history-conditional-reward}, with $f$ being Lipschitz, $\mathrm{P}_X$ satisfying \Cref{def:intrinsic-dimension} with $d_*$, and $f(X)$ satisfying \Cref{def:uc-beta} with exponent $\beta$.
The following theorem parallels \Cref{thm:adaptive-upper,thm:lower-bound} under low-dimensional structure, giving matching upper and lower bounds with $d$ replaced by $d_*$.

\begin{theorem}[Minimax regret under intrinsic dimension]
\label{thm:adaptive-intrinsic}
For every integer $1\leq d_*\leq d$ and every $\beta\in(0,1]$, with candidate-set size $K$ and time horizon $T$, the minimax regret over $\mathcal{F}_{\operatorname{AC}}(d,d_*,\beta)$ satisfies
\begin{align}
\label{eq:adaptive-intrinsic-rate}
\inf_\pi \sup_{\mathrm{P}\in\mathcal{F}_{\operatorname{AC}}(d,d_*,\beta)} \mathbb{E}_{\mathrm{P}}[R(\pi)] \asymp_{\log} (KT)^{\frac{d_*+1-\beta}{d_*+2}} \wedge T^{\frac{d_*+1-\beta}{d_*+2-\beta}}.
\end{align}
The upper bound is attained by the \texttt{(BTN)}\textsuperscript{2} policy \eqref{equ:def-policy-adaptivenn} with $C$ and $\ell=c_0 \cdot \log(KT)$, where $C$ and $c_0$ are sufficiently large constants depending only on the distribution.
\end{theorem}

\subsection{Global Regret and Candidate-Set Size}
\label{sec:global-regret}

Recall in \eqref{equ:generation-selection-gap}, the global regret can be written as the sum of the generation gap $\mathbb{E}\!\left[f^*-\max_{x\in\mathcal{D}} f(x)\right]$ with the global optimum $f^*:=\sup_{x\in\mathcal{X}} f(x)$, and the selection gap investigated under different conditions in Sections~\ref{sec-ambient-dim-regret} and \ref{sec:intrinsic-dimension}.
With details deferred to Section~\ref*{app:derivation-generation-gap}, we can show that under \Cref{def:lower-mass,def:intrinsic-dimension}, the generation gap satisfies $\mathbb{E}\!\left[f^*-\max_{x\in\mathcal{D}} f(x)\right] \lesssim K^{-1/d}$ and $\mathbb{E}\!\left[f^*-\max_{x\in\mathcal{D}}f(x)\right]\lesssim K^{-1/d_*}$, respectively, depending on the ambient or intrinsic dimension.
To have an intuitive understanding of the impact of the candidate-set size $K$ on the global regret, we focus on the ambient-dimension setting (\Cref{def:lower-mass}), noting that the intrinsic-dimension one shares the same pattern.  
It follows from the decomposition \eqref{equ:generation-selection-gap}, Section~\ref*{app:derivation-generation-gap} and \Cref{thm:adaptive-upper}, that the global regret satisfies
\begin{align}
\label{eq:global-regret-rate}
\mathbb{E}\!\left[\sum_{t=1}^T\{f^*-f(X_t)\}\right]
\lesssim_{\log}
TK^{-1/d}
+
(KT)^{\frac{d+1-\beta}{d+2}}
\wedge
T^{\frac{d+1-\beta}{d+2-\beta}}.
\end{align}
The result shows that increasing the candidate-set size $K$ reduces the generation gap and increases the selection gap till it plateaus, together suggesting the following two regimes in terms of $K$.
\begin{itemize}
    \item \textbf{Small $K$.} When $K\lesssim T^{\frac{d}{d+2-\beta}}$, the global regret is dominated by the generation gap and decreases when $K$ grows. Note that this regime covers two sub-regimes in the selection gap with the phase transition at $K \asymp T^{\frac{\beta}{d+2-\beta}}$. %
    \item \textbf{Large $K$.}  When $K\gtrsim T^{\frac{d}{d+2-\beta}}$, the selection gap in particular the term $T^{\frac{d+1-\beta}{d+2-\beta}}$ dominates, suggesting that increasing $K$ provides no improvement or harm in the global regret.
\end{itemize}
{These two regimes, together, suggest that increasing $K$ improves the global-regret rate until the selection gap dominates, after which the rate remains unchanged. In practice, this benefit must be weighed against the cost of generating more candidates, including API queries, token usage, latency, and inference-capacity requirements.

The transition to the plateau occurs at a larger $K$ when $\beta$ is larger, extending the range over which generating more candidates is beneficial. This is consistent with the role of $\beta$: a smaller $\beta$ allows $f(X)$ to confine to a narrow range. When candidates have nearly identical rewards, drawing more candidates offers limited additional utility, even under oracle selection. A larger $\beta$ restricts such concentration, therefore leaving more scope for utility gains from best-of-$K$ selection.}

\section{Collaboration between Users}
\label{sec:collaboration}

With the single-user best-of-$K$ problem studied in \Cref{sec-single-user} as the cornerstone, we next investigate the multiple-user extension formalized as a collaborative bandit problem.

\subsection{Settings}
\label{sec:settings-collaboration}

We extend the setting introduced in \Cref{sec:nonparametric-settings} to  $m$ users instead of one, each observing a candidate set of size $K$ from the same marginal distribution~$\mathrm{P}_X$, while possessing potentially different conditional reward functions $f_i$, $i \in [m]$.
To be specific, for $t \in [T]$, let $\mathcal{F}_t$ be the $\sigma$-field include all users' history up to $t$.  
The candidates $\{\mathcal{D}_{i,t}=\{X_{i,t,j}:j\in[K]\}\}_{i \in [m]}$ are mutually independent and independent of $\mathcal{F}_{t-1}$.  
The nonanticipating policy for user $i \in [m]$ is denoted as $\pi_{i,t}$.
Conditional on $\mathcal{F}_{t-1}$ and
$\{\mathcal{D}_{a,t},X_{a,t}:a\in[m]\}$, the rewards
$\{Y_{i,t}\}_{i\in[m]}$ are mutually independent, take values in $[0,1]$,
and satisfy
\begin{align}
\label{eq:collaborative-history-reward}
\mathbb{E}\!\left[Y_{i,t}\,\middle|\,\mathcal{F}_{t-1},\{\mathcal{D}_{a,t},X_{a,t}:a\in[m]\}\right]
=f_i(X_{i,t}).
\end{align}
The system seeks to minimize the aggregate regret of the joint policy $\pi=\{\pi_i\}_{i=1}^m$,
\begin{align}
    R(\pi) = \sum_{i=1}^m\sum_{t=1}^{T} \mathbb{E}\left[ f_{i}(\pi_i^{*}(\mathcal{D}))-f_{i}(\pi_{i,t}(\mathcal{D})) \mid \mathcal{F}_{t - 1} \right]. \label{equ:def-all-regret}
\end{align}
For each $i \in [m]$, we denote by $\pi_i^{*}(\mathcal{D})\in\argmax_{X\in\mathcal{D}}f_{i}(X)$ the oracle policy with full knowledge of $f_i$ based on the candidate set $\mathcal{D}$.
All the notational details follow \Cref{sec:nonparametric-settings}.
See the collaborative framework illustrated in the right-hand part of \Cref{fig:interaction-process}.
In the above setting, we assume that all users participate in every round.   Appendix~\ref*{app:extensions} further extends  to random arrivals, where each user queries with a given probability at each round.

To minimize the system-level regret with a collection of users, intuitively, one should leverage information from users with similar preference rankings.  
To formalize this intuition, we introduce the cluster structure among users based on their preferred candidates.
Formally, let $r \in [m]$ denote the number of clusters, and assume the existence of a latent partition $\mathcal{C}$ of $[m] = \sqcup_{u \in [r]} \mathcal{C}_u := \sqcup_{u \in [r]}\{i\in[m]:z_i=u\}$, where $z_i\in[r]$ denotes the cluster label of user $i$.
Let $s_u:=|\mathcal{C}_u|$ denote the cluster size. 
For $i, j \in [m]$, $z_i = z_j$ if $\mathrm{D}_{ij,\kappa}^{\mathrm{P}_X} = 0$ for all $\kappa \geq 2$, where $\mathrm{D}_{ij,\kappa}^{\mathrm{P}_X}$ is the
$\kappa$-winner distance (\Cref{def-kappa-winner-distance}), i.e., the total variation distance between the distributions of the best-of-$\kappa$ candidates for users $i$ and $j$. %

Specifically, for an integer $\kappa\ge2$, let $\mathcal{D}=\{X_1,\ldots,X_\kappa\}$ consist of independent draws from $\mathrm{P}_X$. For each user $i\in[m]$, let $Q_{f_i,\kappa}^{\mathrm{P}_X}$ denote the distribution of the oracle winner
\begin{align*}
    X_{f_i}^*(\mathcal{D}) \in \argmax_{x\in\mathcal{D}} f_i(x).
\end{align*}
Write $U_i(x):=\mathrm{P}_X\{f_i(X)\le f_i(x)\}$, which represents the probability that a randomly drawn candidate has reward no larger than that of $x.$ 
If $f_i(X)$ satisfies the AC condition in \Cref{def:uc-beta}, the oracle winner is almost surely unique. %
To characterize its distribution, consider the event that the first candidate $X_1$ is the winner. Conditional on $X_1=x$, this occurs precisely when all the remaining $\kappa-1$ candidates have rewards no larger than $f_i(x)$. By independence, the corresponding probability is $U_i(x)^{\kappa-1}$. Since each of the $\kappa$ candidates is equally likely to win,  we obtain, for every measurable set $A\subseteq\mathcal{X}$,
\begin{align*}
    Q_{f_i,\kappa}^{\mathrm{P}_X}(A) = \mathbb{P}\{X_{f_i}^*(\mathcal{D})\in A\} = \kappa\int_A U_i(x)^{\kappa-1}\,\mathrm{P}_X(dx).
\end{align*}
This representation yields the following distance between winner distributions.

\begin{definition}
\label{def-kappa-winner-distance}
For users $i,j\in[m]$, define
\begin{align}
\mathrm{D}_{ij,\kappa}^{\mathrm{P}_X}
    := \left\|Q_{f_i,\kappa}^{\mathrm{P}_X}-Q_{f_j,\kappa}^{\mathrm{P}_X}\right\|_{\mathrm{TV}} = \frac{\kappa}{2}\int_{\mathcal{X}}\left|U_i(x)^{\kappa-1}-U_j(x)^{\kappa-1}\right|\,\mathrm{P}_X(dx).
    \label{eq:kappa-winner-distance}
\end{align}
For the normalized winner distance  $\mathrm{D}_{ij,\kappa}^{\mathrm{P}_X}/\kappa$, define its multi-scale version 
\begin{align}
    \xi_{ij}(\mathrm{P}_X) := \sup_{\kappa\ge2}\frac{\mathrm{D}_{ij,\kappa}^{\mathrm{P}_X}}{\kappa}, \qquad i,j\in[m].
    \label{eq:pairwise-detection-strength}
\end{align}
\end{definition}

The winner distance in~\eqref{eq:kappa-winner-distance} measures the largest difference between the probabilities that the oracle policies of two users select a candidate from the same measurable subset of $\mathcal{X}$. This distance can be zero even when their reward functions differ, provided that they induce the same preference ordering. 
In fact, if $f_i(X)$ satisfies the AC condition in \Cref{def:uc-beta} for every $i\in\mathcal{C}_u$, zero pairwise winner distances within the cluster imply that there exist functions $\phi_i$ and $r_u$ such that, for $\mathrm{P}_X$-almost every $x\in\mathcal{X}$, %
\begin{align}
    f_i(x)=\phi_i(r_u(x)), \qquad i\in\mathcal{C}_u.
    \label{eq:winner-compatible-reward-structure}
\end{align}
In~\eqref{eq:winner-compatible-reward-structure}, $r_u:\mathcal{X}\to[0,1]$ describes the common preference ranking within cluster $\mathcal{C}_u$, while the strictly increasing, possibly nonlinear maps $\phi_i:[0,1]\to[0,1]$ capture user-specific feedback intensities. 
See Appendix~\ref*{app:derivation-zero-winner-common-trend} for the derivation. 
Representation~\eqref{eq:winner-compatible-reward-structure} and equation \eqref{eq:kappa-winner-distance} characterize a broader class of users who can be clustered together, allowing their data to be pooled even when their reward functions differ. 
This setting includes, as a special case, the identical within-cluster reward functions assumed by classical clustered-bandit methods \citep{gentile2014online, li2019improved}.

\begin{figure}[htbp]
 \centering
\includegraphics[width=\linewidth]{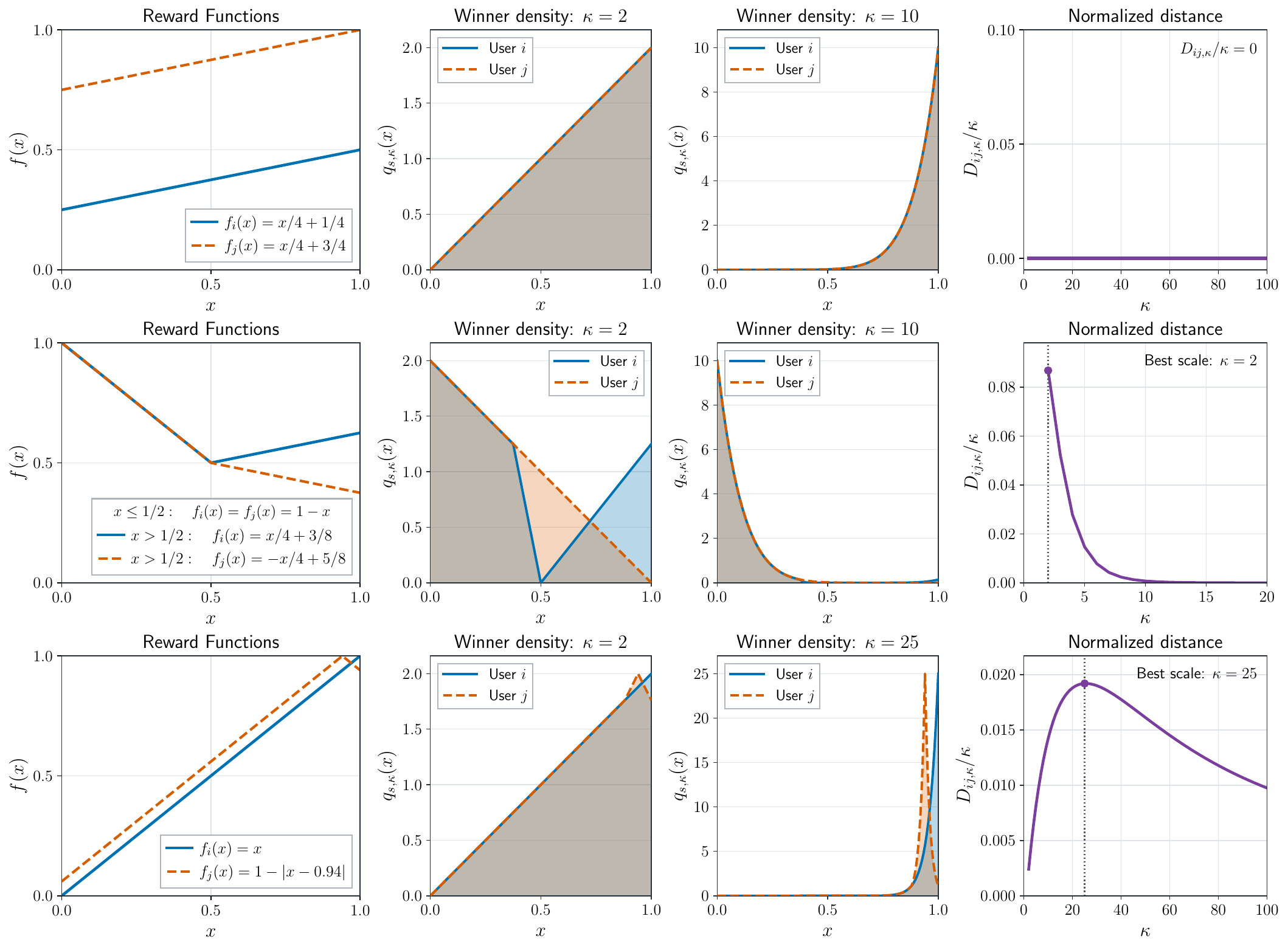}
    \caption{Three example groups of distributions, exhibited as reward functions (left), oracle-winner densities with different $\kappa$ (middle), and normalized winner distances (right) under $\mathrm{P}_X=\operatorname{Unif}([0,1])$.
    The top row shows different reward functions with identical rankings, yielding zero winner distance at every scale. The middle and bottom rows illustrate different patterns of disagreement between winner distributions, with the normalized winner distance attaining its maximum at different values of $\kappa$.}
    \label{fig:multiscale-winner-density-comparison}
\end{figure}

 If $f_i(X)$ and $f_j(X)$ satisfy \Cref{def:uc-beta}, strict monotonicity of $v\mapsto v^{\kappa-1}$ implies that $\mathrm{D}_{ij,\kappa}^{\mathrm{P}_X}$ is either zero for every $\kappa\ge2$ or positive for every $\kappa\ge2$. %
Consequently, at the population level, whether users $i$ and $j$ belong to the same cluster can be determined by checking whether $\mathrm{D}_{ij,\kappa}^{\mathrm{P}_X}$ is zero or positive at any fixed $\kappa\ge2$. However, detecting a positive distance from data can be easier at some (unknown) values of $\kappa$ than at others, as illustrated in Figure~\ref{fig:multiscale-winner-density-comparison}.
We therefore consider the multiscale version of the normalized winner distance defined in~\eqref{eq:pairwise-detection-strength}, which captures the strongest signal across all scales $\kappa\ge2$. 
We will later use $\xi_{ij}(\mathrm{P}_X)$ to formulate a separation condition for exact cluster recovery.

\subsection{Methodology}
\label{sec:collaboration-methodology}

We now develop a collaborative policy with two stages: cluster detection and collaborative learning.
In the first stage, each user performs $T_0$ rounds of uniform exploration. The resulting samples are used to estimate the user-specific reward functions, construct a similarity graph, and cluster the users.
In the second stage, each user plays $T_1$ additional rounds, during which users within the same detected cluster collaborate in estimation and decision-making.
The total horizon is $T = T_0 + T_1$, and we re-index the collaborative rounds by $t \in [T_1]$.
The detected clusters are $\mathcal{F}_{T_0}$-measurable, and the information available at the beginning of collaborative round $t$ is described by $\mathcal{F}_{T_0+t-1}$.

\subsubsection{Stage 1: Cluster Detection}
\label{sec:cluster-detection}

The cluster detection stage consists of two steps: similarity graph construction and clustering.
We first construct a graph on the users, connecting each pair whose estimated detection strength is below a prescribed threshold.
We then delete edges to form a disjoint union of cliques (complete graphs), each of which defines a detected cluster.

\paragraph{Similarity Graph Construction.}

To construct the similarity graph, we first estimate individual reward functions, use these to estimate pairwise winner distances, and then apply a threshold to determine which users are connected.
In particular, at each exploration round $t\in[T_0]$, user $i\in[m]$ observes $\mathcal{D}_{i,t}$, selects $\bar X_{i,t}\sim \operatorname{Unif}(\mathcal{D}_{i,t})$ uniformly at random, and observes the corresponding reward $\bar Y_{i,t}$.
For each user, we apply the $k$-NN estimator \citep{jiang2019non} to obtain the estimator $\widehat{f}_i$.
We also assume access to abundant unlabeled candidate contexts $\widetilde{\mathcal{D}}_n = \{\widetilde{X}_q\}_{q = 1}^n$ from $\mathrm{P}_X$.
The reference sample can be additionally generated cheaply from $\mathrm{P}_X$ or formed by retaining unselected candidates from earlier uniform exploration rounds.
Denote by $\widetilde P_n$ the empirical distribution of $\widetilde{\mathcal{D}}_n$, and define the estimated ranks $\widehat U_i(\widetilde X_q) := \widetilde P_n\bigl(\{x:\widehat f_i(x)\le \widehat f_i(\widetilde X_q)\}\bigr)$.  For any integer $\kappa\ge2$, we estimate the $\kappa$-winner distance by
\begin{align}
    \label{eq:winner-tv-plugin-estimator} \widehat{\mathrm{D}}_{ij,\kappa} := \frac1{2n}\sum_{q=1}^n \left| \kappa\widehat U_i(\widetilde X_q)^{\kappa-1} - \kappa\widehat U_j(\widetilde X_q)^{\kappa-1} \right|
\end{align}
For thresholds $\rho_{ij}>0$ specified later, we define the screened edges by
\begin{align} \label{eq:cluster-edge-screening} 
\widehat E := \left\{(i,j): \sup_{\kappa\ge2}\frac{\widehat{\mathrm{D}}_{ij,\kappa}}{\kappa} \leq \rho_{ij} \right\}.
\end{align}

Although this empirical supremum is taken over infinitely many $\kappa$, Appendix~\ref*{app:adaptive-scale-search} provides an efficient procedure for approximating it to any prescribed accuracy.

We choose $\rho_{ij}$ to cover the estimation error of the normalized winner distance uniformly across scales. 
See \Cref{prop:calibrated-cluster-detection_1st}.
Intuitively, a within-cluster pair has $\mathrm{D}_{ij,\kappa}^{\mathrm{P}_X} / \kappa=0$, so the detector retains its edge whenever $\widehat{\mathrm{D}}_{ij,\kappa}/ \kappa$ estimates $\mathrm{D}_{ij,\kappa}^{\mathrm{P}_X} / \kappa$ with an error less than $\rho_{ij}$.
A between-cluster pair with sufficiently large $\xi_{ij}(\mathrm{P}_X)$ will have a distance exceeding $\rho_{ij}$ at some scale, so its edge is removed.

\paragraph{Clustering.}

Given the similarity graph $\widehat E$ defined in \eqref{eq:cluster-edge-screening}, we retain as many edges as possible while forming a disjoint union of cliques. 
Specifically, choose
\begin{align}    \label{eq:collaboration-cluster-deletion}
    \widehat G\in\argmax_{G'\subseteq\widehat E}\left\{|G'|: ([m],G')\text{ is a disjoint union of cliques}\right\}.
\end{align}
The resulting graph $\widehat G$ is then used in the collaborative phase, with each connected component defining a detected cluster.

The clique constraint is imposed to facilitate the martingale arguments used in analyzing   collaborative learning.
The optimization problem~\eqref{eq:collaboration-cluster-deletion} is known as cluster deletion.  
Although computing an exact solution is NP-hard in general \citep{shamir2004cluster}, scalable approximation algorithms are available \citep{veldt2022correlation,balmaseda2024combinatorial}.
The cluster-detection pseudocode and an illustration of cluster recovery are provided in Appendix~\ref*{sec:experiments-collaboration-recovery} (Algorithm~\ref*{alg:cluster-detection} and Figure~\ref*{fig:collaboration-threshold-graphs}).

\subsubsection{Collaborative Learning}
\label{sec:joint-estimation-decision}

We move on to the collaborative phase with fixed detected clusters, i.e., the connected components of $\widehat G$.
At a high level, we extend the \texttt{(BTN)}\textsuperscript{2} algorithm proposed for the single-user setting in \Cref{sec:adaptive-knn} to a multiple-user version within each detected cluster.

To be specific, let the connected components of $\widehat G$ be $\widehat{\mathcal{C}} = \{ \widehat{\mathcal{C}}_u\}_{u \in [r]}$, with cardinality $\widehat s_u:=|\widehat{\mathcal{C}}_u|$.
Let $\mathcal{Z}_{u,t-1} := \{Z_{i,\tau}:i\in\widehat{\mathcal{C}}_u,\ 1\leq\tau<t\}$ be the collaborative history of the cluster before round $t$, containing $(t-1)\widehat{s}_u$ observations collected only during the collaborative phase. 
The multiple-user-averaged $k$-NN estimator of the reward function for cluster $\widehat{\mathcal{C}}_u$ is therefore
\begin{align}
\label{eq:def-collaborative-estimator}
    \widehat f_{u,k}(x;\mathcal{Z}_{u,t-1})
    :=
    \frac{1}{k}\sum_{q=1}^{k}Y_{i_q,\tau_q}, \qquad k\in[(t-1)\widehat s_u],
\end{align}
where 
$\sigma_q(x;\mathcal{Z}_{u,t-1})=(i_q,\tau_q)\in\widehat{\mathcal{C}}_u\times[t-1]$, $q\in[(t-1)\widehat s_u]$ denotes the (user, round)-index of the $q$th nearest observation to $x$ and  $r_k(x;\mathcal{Z}_{u,t-1}):=\|X_{i_k,\tau_k}-x\|_2$ is the pooled $k$-distance.

Based on $\widehat{f}_{u, k}$, we construct a  confidence bound for each $k$ and then adaptively choose $k$.  
Similarly to \eqref{eq:def-ucb-radius}, for tuning parameters $C>0$ and $\ell>0$,  let $U_{u,k}(x,\mathcal{Z}_{u,t-1}):= C r_k(x;\mathcal{Z}_{u,t-1})+\sqrt{\ell/k}$, and define the UCB score  as
\begin{align*}
    \operatorname{UCB}_{u,t}(x)
    :=
    \widehat f_{u,\widehat k_{u,t}(x)}(x;\mathcal{Z}_{u,t-1})
    +U_{u,\widehat k_{u,t}(x)}(x,\mathcal{Z}_{u,t-1}),
    \qquad t\geq2,
\end{align*}
with $\widehat k_{u,t}(x) := \argmin_{k\in[(t-1)\widehat s_u]}U_{u,k}(x, \mathcal{Z}_{u, t-1})$. 
Then, every user $i\in \widehat{\mathcal{C}}_u$ applies the policy to its own candidate set according to the decision rule 
\begin{align}
\label{eq:def-collaborative-decision}
    X_{i,t}=X_{i,t,J_{i,t}},
    \quad \mbox{with }
J_{i,t}\in\argmax_{j\in[K]}\operatorname{UCB}_{u,t}(X_{i,t,j}),
\end{align}
and then appends $(X_{i,t}, Y_{i, t})$ to its history.

\subsection{Theoretical Results}
\label{subsec:collaboration-theory}

In this subsection, we provide regret guarantees for our two-stage algorithm under known clusters in \Cref{subsubsection-regret-known-partition} and under unknown clusters in \Cref{sec-regret-full-alg}.
Additional theoretical results and discussion are given in Appendix~\ref*{app:extensions}.
To control the concentration of candidates in small neighborhoods, we impose \Cref{ass:upper-mass} below throughout this subsection.

\begin{assumption}
\label{ass:upper-mass}
There exists a constant $\overline{c}>0$ such that $\mathrm{P}_X(B(x,r)) \leq \overline{c}r^d$ for all $x\in\mathcal{X}$ and $r\in(0,1]$.
\end{assumption}

Recall the cluster size $s_u:=|\mathcal{C}_u|$ and define the average reward $\overline f_u(x):=s_u^{-1}\sum_{a\in\mathcal{C}_u}f_a(x)$.
Suppose $\overline f_u(X)$ satisfies \Cref{def:uc-beta} with exponent $\beta_u$ and constant $C_{u,\mathrm{AC}}$.
In Appendix~\ref*{sec:induced-reward-structure}, we prove that such an exponent $\beta_u$ and constant $C_{u,\mathrm{AC}}$ exist and provide explicit choices for both.

\subsubsection{Regret with Oracle Clusters} \label{subsubsection-regret-known-partition}

In this subsection, we assume that the true clusters are known.
Within each cluster~$\mathcal{C}_u$, $u\in[r]$, the collaborative \texttt{(BTN)}\textsuperscript{2} algorithm in \eqref{eq:def-collaborative-decision} pools observations from its members to estimate the average reward $\overline f_u$, which preserves the preference ranking of every user in the cluster.
\Cref{thm:oracle-cluster-regret} below controls the resulting aggregate regret.

\begin{theorem}\label{thm:oracle-cluster-regret}
Suppose every $f_i$, $i \in \mathcal{C}_u$, $u \in [r]$, is $C_L$-Lipschitz, $\overline f_u(X)$ is AC-$\beta_u$ with constant $C_{u,\operatorname{AC}}$, and \Cref{def:lower-mass,ass:upper-mass} hold. 
Run the collaborative \texttt{(BTN)}\textsuperscript{2} algorithm in \eqref{eq:def-collaborative-decision} using the true clusters with $C$ and $\ell:=c_0\log(KmT_1)$, where $C$ and $c_0$ are sufficiently large  constants  depending only on the distribution.
Let $X_{i,t}^{\operatorname{oracle}}$ denote the action selected for user $i$ in round $t$, and define the corresponding regret by
\begin{align*}
    R_i^{\operatorname{oracle}}(T_1)
    := \sum_{t=1}^{T_1}\mathbb{E}\left[\max_{x\in\mathcal{D}_{i,t}}f_i(x)-f_i(X_{i,t}^{\operatorname{oracle}})\right].
\end{align*}
If $s_uK\lesssim T_1^{(d+1-\beta_u)/(d+1)}$, then up to polylogarithmic factors, it holds that 
\begin{align*}
    \sum_{i\in\mathcal{C}_u}R_i^{\operatorname{oracle}}(T_1)\lesssim_{\log} R_u(T_1, K):= (Ks_uT_1)^{\frac{d+1-\beta_u}{d+2}}\wedge\ (s_uT_1)^{\frac{d+1-\beta_u}{d+2-\beta_u}}.
\end{align*}
\end{theorem}

With $T_1$ rounds of interaction, the total sample size within $\mathcal{C}_u$ is $s_uT_1$.
Compared with the single-user result in \Cref{thm:adaptive-upper}, \Cref{thm:oracle-cluster-regret} gives the same rates and phase transition with an effective horizon of $s_uT_1$.
The condition $s_uK\lesssim T_1^{(d+1-\beta_u)/(d+1)}$ requires the cluster size and candidate-set size to remain moderate relative to the horizon.
This condition ensures that the leading selection term absorbs the cost of simultaneous decisions within each cluster.
For larger clusters, Appendix~\ref*{app:collaboration-theory-details} provides the complete regret bound and analysis without the restriction $s_u\lesssim K^{-1}T_1^{(d+1-\beta_u)/(d+1)}$. %
The proof of the upper bound is provided in Appendix~\ref*{app:clusterwise-pooling-proofs}.
A matching lower bound is established in \Cref{thm:oracle-cluster-lower}, %
showing that the rate is minimax rate-optimal up to logarithmic factors.

\begingroup

\begin{theorem}
\label{thm:oracle-cluster-lower}
Let $\mathcal{F}_{\operatorname{col}}(d,\beta_u,s_u)$ denote the class of $d$-dimensional distributions satisfying the conditions in \Cref{sec:settings-collaboration} and \Cref{thm:oracle-cluster-regret}, with $s_u$ users forming each cluster.
For any nonanticipating joint policy $\pi$ that knows the true cluster $\mathcal{C}$, let $X_{i,t}^{\operatorname{oracle}}$ denote the action selected under $\pi$ and write $R_i^{\operatorname{oracle}}(\pi,T_1)$ for the regret defined in \Cref{thm:oracle-cluster-regret}. 
For every integer $d\geq1$ and $\beta_u\in(0,1]$, for all $K\geq2$ and all sufficiently large $s_uT_1$, we have
\begin{align*}
\inf_{\pi}\sup_{\mathrm{P}\in\mathcal{F}_{\operatorname{col}}(d,\beta_u,s_u)}\sum_{i\in\mathcal{C}_u}R_i^{\operatorname{oracle}}(\pi,T_1)\gtrsim (Ks_uT_1)^{\frac{d+1-\beta_u}{d+2}}\wedge(s_uT_1)^{\frac{d+1-\beta_u}{d+2-\beta_u}}.
\end{align*}
\end{theorem}

\endgroup

To further understand the collaboration gains, we compare with separate learning where $s_u$ users run \texttt{(BTN)}\textsuperscript{2} based on their own history.
Suppose that each $f_i(X)$ also satisfies AC-$\beta_u$.  
By \Cref{thm:adaptive-upper}, the aggregate regret of separate learning is $\{(KT)^{\frac{d+1-\beta}{d+2}}
\wedge
T^{\frac{d+1-\beta}{d+2-\beta}}\}s_u$.
The collaborative rule improves by factors of $s_u^{(1+\beta_u)/(d+2)}$ and $s_u^{1/(d+2-\beta_u)}$ when both bounds are unsaturated and when both are saturated, respectively.
In the intermediate region, only the separate learning bound has saturated, and the gain decreases between these two factors as $K$ increases.
Overall, collaboration reduces the aggregate regret, with gains growing both in the cluster size $s_u$ and in  $\beta_u$, indicating that information sharing is more beneficial when reward functions are less concentrated and easier to distinguish.

Although \Cref{thm:oracle-cluster-regret} guarantees an improvement in the aggregate regret bound, it does not by itself guarantee an improvement for every user.
Appendix~\ref*{sec:oracle-individual-regret} characterizes individual regret under collaboration.
We show that users with stronger reward signals bear larger shares of the collaborative regret, and that collaborating with weaker-signal users may yield a worse individual regret guarantee than separate learning.

As for the global regret \eqref{equ:generation-selection-gap}, it additionally includes the generation gap of order $s_u T_1 K^{-1/d}$, as discussed in \Cref{sec:global-regret}. 
The global regret bound continues to decrease at this rate until $K\asymp(s_uT_1)^{d/(d+2-\beta_u)}$, beyond which it remains of order $(s_uT_1)^{(d+1-\beta_u)/(d+2-\beta_u)}$. 
Collaboration therefore extends the range over which increasing~$K$ remains useful by a factor of $s_u^{d/(d+2-\beta_u)}$, broadening the scope of effective best-of-$K$ selection.
See \Cref*{sec:oracle-global-regret} for more details.

\subsubsection{Regret with Unknown Clusters} \label{sec-regret-full-alg}

When the oracle clusters are unknown, we first estimate the clusters and then run the collaborative \texttt{(BTN)}\textsuperscript{2} algorithm within each estimated cluster.  Before establishing a regret bound for this two-stage procedure, we provide a sufficient condition for exact recovery of the true clusters in \Cref{prop:calibrated-cluster-detection_1st}.  For any $\delta \in (0, 1)$, define the noise level
\begin{align}\label{equ:def-of-en}
    e_n:=\left\{\frac{\log^2(4mT_0^2/\delta)}{T_0}\right\}^{\beta_i/(d+2)}+\left\{\frac{\log^2(4mT_0^2/\delta)}{T_0}\right\}^{\beta_j/(d+2)}+\sqrt{\frac{\log^2\{8m^2n/\delta\}}{n}}.
\end{align}
With probability at least $1-\delta$, this quantity bounds the normalized estimation error $\kappa^{-1}|\widehat{\mathrm{D}}_{ij,\kappa}-\mathrm{D}_{ij,\kappa}^{\mathrm{P}_X}|$ simultaneously over all user pairs and all $\kappa\ge2$.
The first two terms arise from estimating $f_i$ and $f_j$ using their labeled exploration samples, while the third controls the sampling error from the shared unlabeled sample.
Since unlabeled samples are inexpensive, $n$ can be chosen sufficiently large for the third term to be absorbed.

\begin{proposition}[Consistent cluster detection]
\label{prop:calibrated-cluster-detection_1st}
Fix $\delta\in(0,1)$. 
Suppose each $f_i$ is $C_L$-Lipschitz with $C_L>0$ an absolute constant, each $f_i(X)$ satisfies AC-$\beta_i$ (\Cref{def:uc-beta}) with $\beta_i>0$, and $\mathrm{P}_X$ satisfies \Cref{def:lower-mass}.
Construct the estimates of each $f_i$ using suitably tuned $k$-NN estimators, as detailed in Appendix~\ref*{app:cluster-detection-proofs}.
For sufficiently large $T_0$ and $n$, suppose $\rho_{ij}\ge e_n$ for every $i<j$, and $\xi_{ij}(\mathrm{P}_X)>\rho_{ij}+e_n$ for every between-cluster pair. 
Then the screening rule~\eqref{eq:cluster-edge-screening} followed by the clustering satisfies $\Pr\!\left(\widehat{\mathcal{C}}=\mathcal{C}^*\right)\ge1-\delta$, where $\widehat{\mathcal{C}}$ and $\mathcal{C}^*$ are the estimated and true clusters.
\end{proposition}

\Cref{prop:calibrated-cluster-detection_1st} provides sufficient conditions for exact recovery of the true clusters with probability at least $1-\delta$.
First, every between-cluster pair is well-separated with a large enough winner distance $\xi_{ij}(\mathrm{P}_X)>2e_n$.
Second, the thresholds are properly tuned such that $\rho_{ij}\ge e_n$ for all pairs $i<j$ and lie in $[e_n,\xi_{ij}(\mathrm{P}_X)-e_n)$ for between-cluster pairs.
In this case, for a within-cluster pair, the population signal is zero, so its empirical supremum is at most $e_n\le\rho_{ij}$ and the edge is retained.
For a between-cluster pair, the lower bound $\xi_{ij}(\mathrm{P}_X)-e_n>\rho_{ij}$ ensures that its edge is deleted.
With these conditions, the screened graph from the first step is already a disjoint union of  cliques and satisfies the constraint in \eqref{eq:collaboration-cluster-deletion}, so the clustering step leaves it unchanged as $\widehat G=\widehat E$.

We now focus on the regret guarantee under exact cluster recovery.

\begin{theorem}
\label{thm:clique-imperfect-regret}
Suppose the conditions of \Cref{thm:oracle-cluster-regret} and \Cref{prop:calibrated-cluster-detection_1st} hold with $\delta=T^{-3}$.
Let $\pi_{\mathtt{(BTN)}^2}^{\mathrm{col}}$ be the collaborative \texttt{(BTN)}\textsuperscript{2} policy in \eqref{eq:def-collaborative-decision} with $C$ and $\ell:=c_0\log(KmT_1)$, where $C$ and $c_0$ are sufficiently large numerical constants only depending on the distribution.
It then holds
\begingroup
\begin{align*}
    \mathbb{E}\left[R(\pi_{\mathtt{(BTN)}^2}^{\mathrm{col}})\right] \lesssim_{\log} mT_0 + \sum_{u=1}^{r}R_u(T_1, K),
\end{align*}
\endgroup
where $R_u(T_1,K)$ is the oracle-cluster bound in \Cref{thm:oracle-cluster-regret}.
\end{theorem}

\Cref{thm:clique-imperfect-regret} shows that, under exact cluster recovery, collaborative \texttt{(BTN)}\textsuperscript{2} attains the oracle collaboration bound up to an additional cost $mT_0$.
This cost arises from the initial $T_0$ exploration rounds undertaken by the $m$ users.
When exact recovery is guaranteed by a small exploration budget such that $mT_0\lesssim\sum_{u=1}^{r}R_u(T_1,K)$, the regret bound inherits the rate established in \Cref{thm:oracle-cluster-regret}, along with the collaboration gains discussed after that theorem.
The required exploration horizon depends on the unknown separation condition between clusters, with stronger separation permitting a smaller $T_0$.
In fact, with a mild separation condition, the exploration cost is negligible for a feasible choice of $T_0$.

\begin{corollary} 
\label{cor:etc-fixed-separation}
Suppose the conditions in \Cref{thm:clique-imperfect-regret} hold. 
Assume that there exists a constant $\xi_0>0$ such that $\xi_{ij}(\mathrm{P}_X)\geq\xi_0$ whenever $z_i\ne z_j$.
For a fixed problem instance and all sufficiently large $T$, choosing $T_0=\lceil T^{d/(d+2)}\rceil$, the collaborative \texttt{(BTN)}\textsuperscript{2} attains
\begin{align*}
\mathbb{E}[R(\pi_{\mathtt{(BTN)}^2}^{\mathrm{col}})]\lesssim_{\log}\sum_{u=1}^{r}R_u(T,K).
\end{align*}
\end{corollary}

In \Cref{cor:etc-fixed-separation}, choosing $T_0\asymp T^\eta$ for any fixed $0<\eta<d/(d+2)$ in \Cref{thm:clique-imperfect-regret}, \texttt{(BTN)}\textsuperscript{2} ensures an exact cluster recovery and the exploration cost is dominated.  This implies that in practice,  $T_0$ can be prescribed from~$T$ without specific tuning. %
When $T$ is unknown, the same regret order follows by applying the doubling trick \citep[e.g.][]{auer1995gambling,zhao2026contextual}.  To relax the separation condition, we consider a weak separation condition in \Cref*{app:weak-separation}, which permits regret guarantees without exact cluster recovery.

\section{Experiments}
\label{sec:experiments}

We validate our theory through simulations in Section~\ref{sec:experiments-simulation}.
We then evaluate our method on a real question-answering benchmark with constructed user preferences in \Cref{sec:experiments-real-data}.

\subsection{Simulation}
\label{sec:experiments-simulation}

We first study single-user selection without and with low-dimensional structure in Sections~\ref{sec:experiments-upper-tail-scaling} and~\ref{sec:experiments-intrinsic-dimension}.
We then examine cluster detection and collaboration in Sections~\ref{sec:experiments-cluster-recovery} and~\ref{sec:experiments-oracle-collaboration}, respectively.
See all additional results, implementation and dataset details in Appendix~\ref*{app:knn-experiments}. 
In simulation, the underlying reward function $f$ is known, and we report the candidate-set regret \eqref{equ:def-regret-single}.
We also report comparisons of global regret \eqref{eq:global-regret-rate} in the appendix.
For each setting in this subsection, we report means over 100 independent Monte Carlo repetitions with pointwise 95\% Student-$t$ confidence intervals.

\subsubsection{Single-User Selection}
\label{sec:experiments-upper-tail-scaling}

At round $t$, the candidate set contains $K$ independent draws from $\operatorname{Unif}[0,1]$. For $\beta\in\{0.5,0.75,1\}$, the conditional reward mean and feedback law are
\begin{align}
    f_\beta(x)=1-(1-x)^{1/\beta} \quad \mbox{and} \quad Y_t\mid X_t=x\sim\operatorname{Bernoulli}\{f_\beta(x)\}.
    \label{eq:simulation-reward-model}
\end{align}
We can verify the family satisfies the conditions used in \Cref{thm:adaptive-upper} with AC-$\beta$.

\begin{figure}[htbp]
    \centering
    \includegraphics[width=\textwidth]{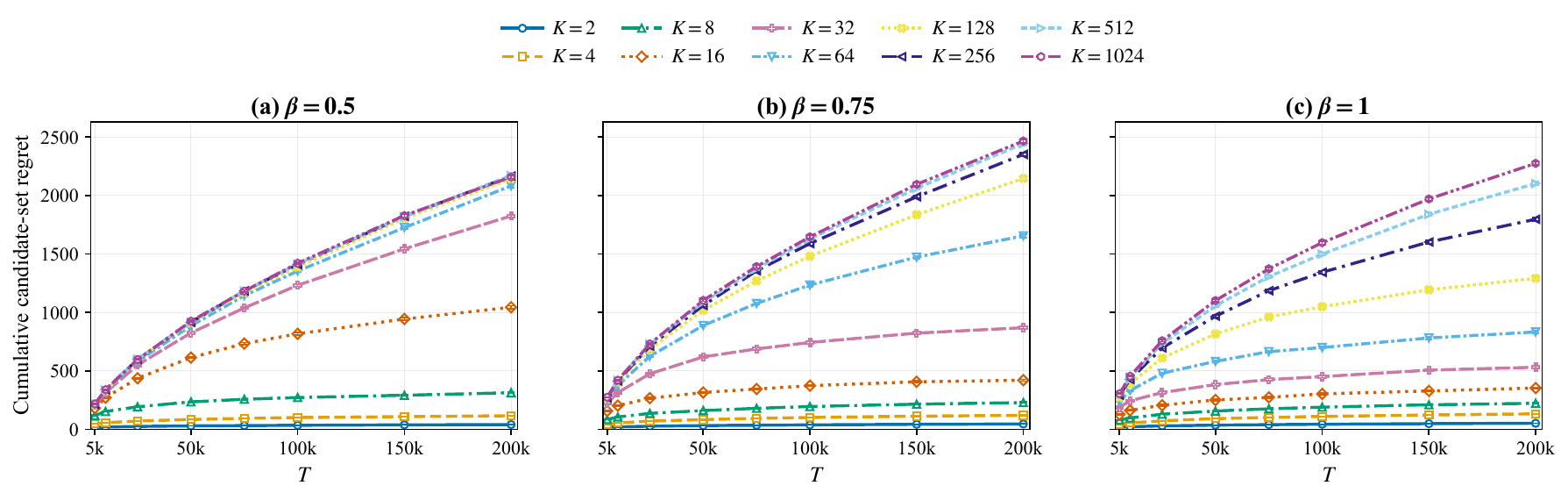}
    \caption{Single-user candidate-set regret across different horizons $T$.
    Each curve corresponds to a candidate-set size $2\leq K\leq1024$.}
    \label{fig:exp47-regret-vs-t}
\end{figure}

We first examine the regret of \texttt{(BTN)}\textsuperscript{2} as $T$ varies in Figure~\ref{fig:exp47-regret-vs-t}. 
Within each panel, regret increases with $T$ for every displayed candidate-set size.
At a fixed horizon, increasing $K$ raises regret in the small-$K$ range, whereas the curves become close once~$K$ is sufficiently large. This saturation, clearest for $\beta=0.5$, agrees with the two-phase rate $(KT)^{\frac{d+1-\beta}{d+2}}\wedge T^{\frac{d+1-\beta}{d+2-\beta}}$ in~\eqref{eq:adaptive-rate}.
The first term increases with both $T$ and $K$, while the second continues to grow with $T$ but no longer depends on $K$. The growth with $T$ also becomes slower as $\beta$ increases at the largest tested $K$, consistent with the decreasing horizon exponents in this rate.
At fixed $T$, increasing $\beta$ shifts the transition to larger $K$, so a larger candidate set is needed to reach the plateau. This agrees with the transition scale $K\asymp T^{\beta/(d+2-\beta)}$ obtained by balancing the two terms in~\eqref{eq:adaptive-rate}.
Appendix~\ref*{app:candidate-set-transition} provides complementary $K$--regret curves.

\subsubsection{Distribution with Intrinsic Dimensions}
\label{sec:experiments-intrinsic-dimension}

\begin{figure}[htbp]
    \centering
    \includegraphics[width=\textwidth]{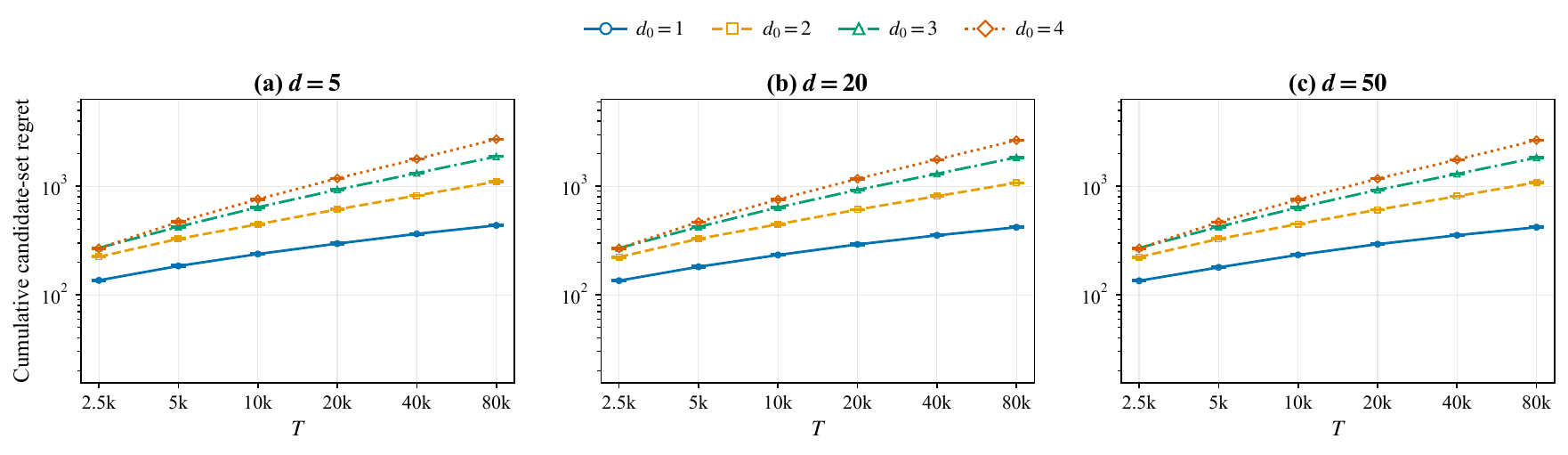}
    \caption{Single-user candidate-set regret under different ambient dimensions, with $K=8$ and $\tau=0.05$.
    Panels correspond to $d\in\{5,20,50\}$, and curves vary $d_0\in\{1,2,3,4\}$.}
    \label{fig:exp45-ambient-dimension}
\end{figure}

We next examine distributions with low-dimensional structure. 
Let $d_0$ denote the signal dimension, with $1\leq d_0<d$, and draw $U\sim\operatorname{Unif}([0,1 / \sqrt{d_0}]^{d_0})$ and $V\sim\operatorname{Unif}(B_2^{d-d_0}(0,\tau))$ independently, where $B_2^m(0,\tau)=\{v\in\mathbb{R}^m:\lVert v\rVert_2\leq\tau\}$ and $\tau$ is a small radius.
The observed context is $Z=(U, V)$. 
The learner observes $Z$ and receives Bernoulli feedback with mean $f(Z) = f(U, V)=1-\lVert\mathbf{1} /\sqrt{d_0}-U\rVert_2$. 
The marginal distribution is a small orthogonal perturbation, of radius $\tau$, of a distribution supported on a $d_0$-dimensional compact manifold, so $d_0$ controls the dimension of the underlying low-dimensional structure.
Figure~\ref{fig:exp45-ambient-dimension} compares candidate-set regret across ambient dimensions $d\in\{5,20,50\}$. Within each ambient dimension, regret grows more quickly with the horizon as the signal dimension $d_0$ increases. For each signal dimension, the curves remain nearly identical across the three ambient dimensions: at every matched $(d_0,T)$, the largest mean regret is at most $1.041$ times the smallest. These results illustrate little sensitivity to ambient dimension in this setting.

\subsubsection{Cluster Recovery}
\label{sec:experiments-cluster-recovery}

\begin{figure}[htbp]
    \centering
    \includegraphics[width=\textwidth]{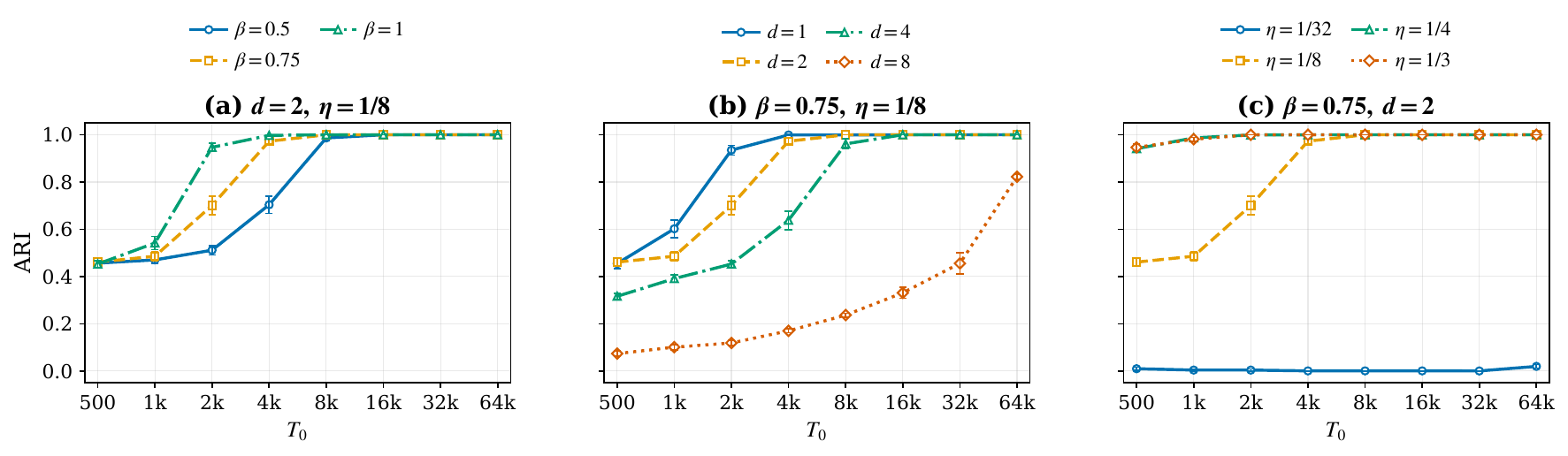}
    \caption{ARI versus exploration size $T_0$, with different $\beta$, $d$, and $\eta$. }
    \label{fig:exp137-cluster-recovery}
\end{figure}

We investigate the cluster-detection procedure. 
We simulate 18 users divided equally into three clusters, with $X\sim\operatorname{Unif}([0,1]^d)$.
Reusing $f_\beta$ from~\eqref{eq:simulation-reward-model}, user $i$ in group $z_i\in\{1,2,3\}$ has reward mean
\begin{align*}
    f_i(x)=\frac{1-c_i}{2}+c_i f_\beta\!\left(1-2\min\left\{|x_1-(z_i-1)\eta|,\,1-|x_1-(z_i-1)\eta|\right\}\right).
\end{align*}
Each group contains the six scales $c_i\in\{0.3,0.4,0.5,0.6,0.7,0.8\}$, and feedback is Bernoulli with mean $f_i(x)$. 
The reward functions have period one and shifts $0,\eta,2\eta$ across the three groups, with greater separation for larger $\eta\in(0,1/3]$.
With $T_0$ observations per user, we apply the cluster-detection procedure in Section~\ref{sec:cluster-detection}, estimating rewards by $k$-nearest-neighbor regression with the rounded and clipped neighbor count specified in Appendix~\ref*{app:experiment-details}. We set $\delta=0.05$ and $\rho_{ij}=e_n/16$, with $e_n$ defined in~\eqref{equ:def-of-en}.
We measure recovery performance using the adjusted Rand index (ARI).
Figure~\ref{fig:exp137-cluster-recovery} validates \Cref{prop:calibrated-cluster-detection_1st} by showing that recovery generally improves with larger $T_0$ and $\beta$, smaller $d$, and greater cluster separation. 
Well-separated clusters can be accurately recovered with relatively few explorations: when $\eta=1/4$, mean ARI exceeds $0.94$ with only $T_0=500$ observations per user.

\subsubsection{Collaborative Learning within an Oracle Cluster}
\label{sec:experiments-oracle-collaboration}

\begin{figure}[htbp]
    \centering
    \includegraphics[width=\textwidth]{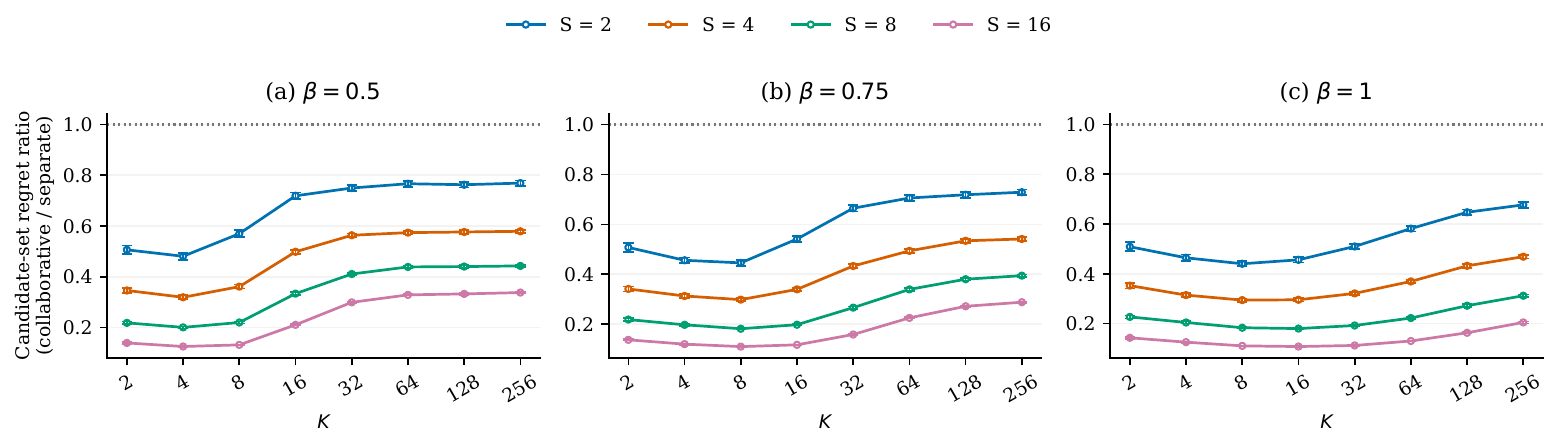}
    \caption{Aggregate candidate-set regret ratios for collaborative versus separate learning within an oracle cluster.
    Panels vary $\beta$, and curves vary $S$. 
    Ratios are computed within each repetition and then averaged.
    The dotted line marks equal aggregate regret.}
    \label{fig:exp56-oracle-collaboration}
\end{figure}

We compare collaborative and separate \texttt{(BTN)}\textsuperscript{2} learning within one known cluster of $S$ users.
Using $f_\beta$ from~\eqref{eq:simulation-reward-model}, user $i$ has reward mean $f_i(x)=(1-c_i)/2+c_i f_\beta(x)$, where $c_i=0.3+0.6(i-1)/(S-1)$.
Thus, the users have zero pairwise winner distances. 
Each user interacts $T=80000$ rounds, with $S\in \{2,4,8,16\}$, $\beta\in\{0.5, 0.75,1\}$, and $K$ varying from $2$ to $256$. 
Figure~\ref{fig:exp56-oracle-collaboration} reports the ratio of aggregate regret under collaborative learning to that under separate learning. 
The mean ratios are below one across all settings and decrease with cluster size, indicating greater gains from collaboration in larger number of users.
The results show benefits of collaboration despite differences in the reward function values, supporting our preference-based cluster definition.

\subsection{Real Data Experiments}
\label{sec:experiments-real-data}

We evaluate our methods on a question-answer benchmark against other baselines.

\subsubsection{Data and Baselines}
\begin{figure}[htbp]
    \centering
    \includegraphics[width=\textwidth]{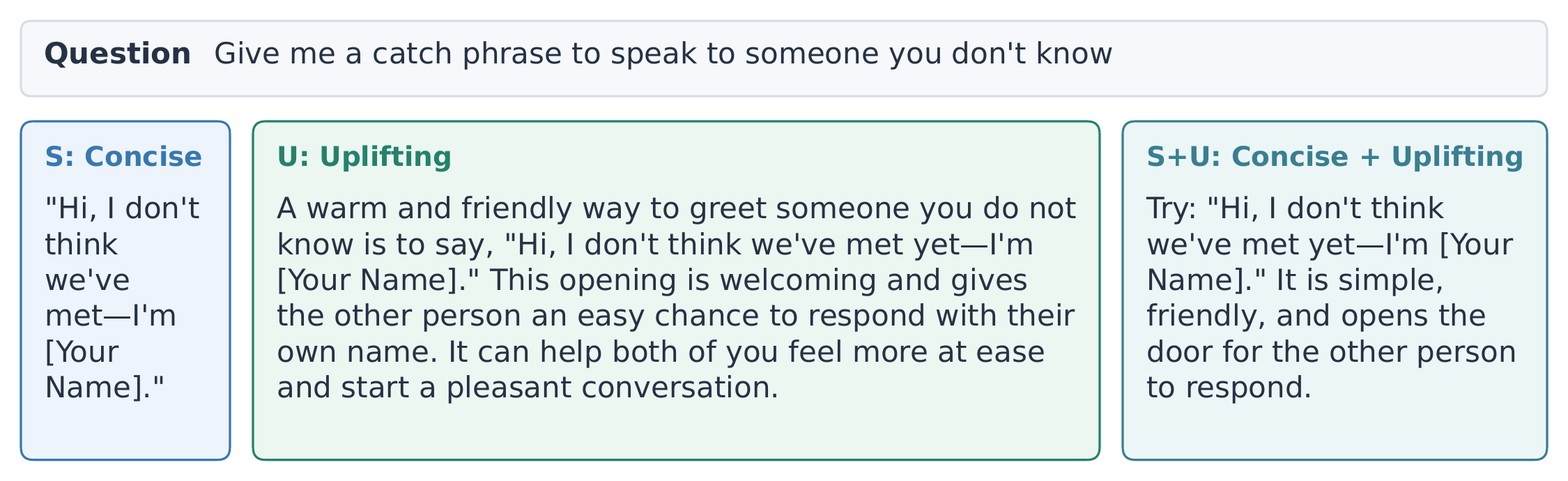}
    \caption{Example of responses to the same question with different styles: concise (S), uplifting (U), and their combination (S+U).}
    \label{fig:realdata-style-example}
\end{figure}

We use 2,000 self-contained questions from LMSYS-Chat-1M \citep{zheng2024lmsys}.
We consider four response styles: creative (C), verbose (V), concise (S), and uplifting (U), and their six pairwise combinations (C+V, C+S, etc.), with examples shown in Figure~\ref{fig:realdata-style-example}.
For each question, we generate 11 answers: one neutral answer and one in each of these ten styles.
The full pipeline for question selection and answer generation is detailed in Appendix~\ref*{app:real-data-details}.
To generate the context vector, we use a 256-dimensional embedding \citep{alibabacloud2026text}.
For a question $q$ and one of its answers $a$, we define the context $x_{q,a}=\operatorname{embd}(q,a)-\operatorname{embd}(q,a_{q,0})$, where $a_{q,0}$ is the neutral answer to $q$.

For fixed $K\in\{2,4,6,8,10\}$, we form the candidate set at each interaction from the context vectors of $K$ distinct answers sampled uniformly from the 11 generated for question $q$.
The policy selects one answer for the user and observes its reward.
Eight users form two clusters of four.
The reward is the deterministic value in Table~\ref{tab:realdata-preferences} for the user's cluster and the selected answer's category.
We report cumulative global regret per user relative to the maximum reward among the 11 answer categories, namely $1$.

\begin{table}[htbp]
\centering
\makeatletter
\let\realdata@makecaption\@makecaption
\long\def\@makecaption#1#2{\begingroup\centering\realdata@makecaption{#1}{#2}\endgroup}
\makeatother
\caption{Constructed reward scores for the two user clusters.}
\label{tab:realdata-preferences}
\begin{tabular}{lrrrrrrrrrrr}
\toprule
Cluster & N & C & V & S & U & CV & CS & CU & VS & VU & SU \\
\midrule
G1 & $.5$ & $1$ & $-1$ & $-1$ & $-1$ & $-1$ & $-1$ & $-1$ & $1$ & $1$ & $-1$ \\
G2 & $.5$ & $1$ & $-1$ & $-1$ & $1$ & $-1$ & $-1$ & $-1$ & $-1$ & $-1$ & $-1$ \\
\bottomrule
\end{tabular}
\end{table}

For separate learning, we compare with linear stochastic bandit methods \citep{li2010contextual, agrawal2013thompson}.
Under oracle collaboration, we use the same baselines with observations pooled within the true clusters.
With unknown clusters, we compare with methods from the clustering bandit literature \citep{gentile2014online, li2019improved, gentile2017on, li2016collaborative, wang2023online, li2025demystifying, ban2024meta}, including the variants described in these papers.
For \texttt{(BTN)}\textsuperscript{2}, after initial exploration, we periodically update the estimated clusters and use their pooled feedback for response selection.
Implementation details are provided in Appendix~\ref*{app:realdata-methods}.

\subsubsection{Results}
\label{sec:realdata-independent-oracle}
\label{sec:realdata-unknown}

All main comparisons of global regret are reported in Table~\ref{tab:realdata-v4-two-comparison}.
\texttt{(BTN)}\textsuperscript{2} achieves the lowest mean global regret among the compared methods in every reported setting.
Increasing $K$ lowers the global regret of \texttt{(BTN)}\textsuperscript{2} and strengthens its relative advantage over linear methods under separate learning.
Collaboration further reduces the regret of \texttt{(BTN)}\textsuperscript{2} at every tested $K$, and its mean regret with unknown clusters is at most $3\%$ above the oracle cluster case.
Thus, the results illustrate the effectiveness of collaboration and the designed cluster detection procedure.

\begin{table}[htbp]
\centering
\vspace{4pt}
\caption{
Global regret under different settings on the real-data benchmark.
\textbf{Bold} marks the lowest mean for each setting.
\colorbox[HTML]{DDE5DE}{Pale green} highlights results that significantly outperform all competing methods in Wilcoxon tests at the $.05$ level. }
\label{tab:realdata-v4-two-comparison}
\label{tab:realdata-v4-two-unknown}
\begingroup
\definecolor{realdataSage}{HTML}{DDE5DE}
\footnotesize\setstretch{1}\fontsize{8.5pt}{10pt}\selectfont\setlength{\tabcolsep}{3.1pt}
\renewcommand{\arraystretch}{1.05}
\begin{tabular}{llccccc}
\toprule
Mode & Method & $K=2$ & $K=4$ & $K=6$ & $K=8$ & $K=10$ \\
\midrule
\multicolumn{7}{l}{\textbf{(a) Cluster G1}} \\
\midrule
 & \texttt{Random} & $2633.3 \pm 25.3$ & $2633.5 \pm 27.0$ & $2632.1 \pm 21.9$ & $2633.1 \pm 23.8$ & $2635.1 \pm 20.0$ \\
\midrule
\multirow{3}{*}{\shortstack[l]{Separate\\Learning}} & \underline{\texttt{(BTN)}\textsuperscript{2}} & \cellcolor{realdataSage}$\mathbf{2271.6 \pm 41.6}$ & \cellcolor{realdataSage}$\mathbf{1928.0 \pm 37.2}$ & \cellcolor{realdataSage}$\mathbf{1656.2 \pm 25.8}$ & \cellcolor{realdataSage}$\mathbf{1419.6 \pm 28.1}$ & \cellcolor{realdataSage}$\mathbf{1186.8 \pm 11.9}$ \\
 & \texttt{LinUCB} & $2382.7 \pm 29.5$ & $2156.5 \pm 28.9$ & $1940.4 \pm 34.0$ & $1735.5 \pm 86.7$ & $1523.4 \pm 99.8$ \\
 & \texttt{LinTS} & $2380.1 \pm 27.6$ & $2155.0 \pm 26.4$ & $1946.0 \pm 32.9$ & $1733.1 \pm 39.1$ & $1507.7 \pm 30.1$ \\
\cmidrule(lr){1-7}
\multirow{3}{*}{\shortstack[l]{Oracle\\Clusters}} & \underline{\texttt{(BTN)}\textsuperscript{2}} & \cellcolor{realdataSage}$\mathbf{2054.1 \pm 30.1}$ & \cellcolor{realdataSage}$\mathbf{1617.4 \pm 23.1}$ & \cellcolor{realdataSage}$\mathbf{1460.6 \pm 16.4}$ & \cellcolor{realdataSage}$\mathbf{1312.8 \pm 17.3}$ & \cellcolor{realdataSage}$\mathbf{1148.1 \pm 14.1}$ \\
 & \texttt{LinUCB} & $2342.6 \pm 29.3$ & $2047.7 \pm 31.9$ & $1800.6 \pm 37.0$ & $1550.3 \pm 30.4$ & $1319.8 \pm 23.2$ \\
 & \texttt{LinTS} & $2343.0 \pm 29.6$ & $2043.9 \pm 31.8$ & $1791.3 \pm 33.2$ & $1550.3 \pm 25.3$ & $1315.1 \pm 21.4$ \\
\midrule
\multicolumn{7}{l}{\textbf{(b) Cluster G2}} \\
\midrule
 & \texttt{Random} & $2998.8 \pm 26.0$ & $3001.7 \pm 22.8$ & $3002.1 \pm 20.8$ & $3000.1 \pm 17.6$ & $2999.2 \pm 16.8$ \\
\midrule
\multirow{3}{*}{\shortstack[l]{Separate\\Learning}} & \underline{\texttt{(BTN)}\textsuperscript{2}} & \cellcolor{realdataSage}$\mathbf{2567.6 \pm 26.6}$ & \cellcolor{realdataSage}$\mathbf{2108.6 \pm 38.1}$ & \cellcolor{realdataSage}$\mathbf{1761.3 \pm 26.7}$ & \cellcolor{realdataSage}$\mathbf{1484.6 \pm 21.4}$ & \cellcolor{realdataSage}$\mathbf{1227.2 \pm 43.0}$ \\
 & \texttt{LinUCB} & $2744.5 \pm 38.4$ & $2467.4 \pm 54.6$ & $2205.9 \pm 39.3$ & $1914.8 \pm 23.3$ & $1621.0 \pm 36.2$ \\
 & \texttt{LinTS} & $2742.0 \pm 34.0$ & $2474.0 \pm 34.8$ & $2203.1 \pm 29.4$ & $1918.9 \pm 32.6$ & $1614.1 \pm 22.7$ \\
\cmidrule(lr){1-7}
\multirow{3}{*}{\shortstack[l]{Oracle\\Clusters}} & \underline{\texttt{(BTN)}\textsuperscript{2}} & \cellcolor{realdataSage}$\mathbf{2454.3 \pm 29.9}$ & \cellcolor{realdataSage}$\mathbf{1875.0 \pm 25.5}$ & \cellcolor{realdataSage}$\mathbf{1597.9 \pm 19.3}$ & \cellcolor{realdataSage}$\mathbf{1389.5 \pm 25.0}$ & \cellcolor{realdataSage}$\mathbf{1187.4 \pm 8.5}$ \\
 & \texttt{LinUCB} & $2690.0 \pm 36.2$ & $2326.3 \pm 38.8$ & $2013.7 \pm 39.6$ & $1702.9 \pm 31.5$ & $1372.4 \pm 17.5$ \\
 & \texttt{LinTS} & $2687.7 \pm 38.5$ & $2319.4 \pm 45.8$ & $2010.2 \pm 39.9$ & $1696.2 \pm 32.1$ & $1371.4 \pm 15.2$ \\
\midrule
\multicolumn{7}{l}{\textbf{(c) All Users}} \\
\midrule
 & \texttt{Random} & $2816.1 \pm 17.2$ & $2817.6 \pm 17.8$ & $2817.1 \pm 15.8$ & $2816.6 \pm 15.6$ & $2817.2 \pm 13.4$ \\
\midrule
\multirow{3}{*}{\shortstack[l]{Separate\\Learning}} & \underline{\texttt{(BTN)}\textsuperscript{2}} & \cellcolor{realdataSage}$\mathbf{2419.6 \pm 25.9}$ & \cellcolor{realdataSage}$\mathbf{2018.3 \pm 27.9}$ & \cellcolor{realdataSage}$\mathbf{1708.8 \pm 18.9}$ & \cellcolor{realdataSage}$\mathbf{1452.1 \pm 18.0}$ & \cellcolor{realdataSage}$\mathbf{1207.0 \pm 23.2}$ \\
 & \texttt{LinUCB} & $2563.6 \pm 25.5$ & $2312.0 \pm 33.4$ & $2073.1 \pm 27.7$ & $1825.1 \pm 45.3$ & $1572.2 \pm 54.2$ \\
 & \texttt{LinTS} & $2561.1 \pm 22.7$ & $2314.5 \pm 24.3$ & $2074.5 \pm 23.4$ & $1826.0 \pm 27.1$ & $1560.9 \pm 19.0$ \\
\cmidrule(lr){1-7}
\multirow{3}{*}{\shortstack[l]{Oracle\\Clusters}} & \underline{\texttt{(BTN)}\textsuperscript{2}} & \cellcolor{realdataSage}$\mathbf{2254.2 \pm 22.2}$ & \cellcolor{realdataSage}$\mathbf{1746.2 \pm 18.5}$ & \cellcolor{realdataSage}$\mathbf{1529.2 \pm 14.9}$ & \cellcolor{realdataSage}$\mathbf{1351.2 \pm 15.7}$ & \cellcolor{realdataSage}$\mathbf{1167.7 \pm 8.2}$ \\
 & \texttt{LinUCB} & $2516.3 \pm 25.4$ & $2187.0 \pm 29.1$ & $1907.1 \pm 31.4$ & $1626.6 \pm 23.4$ & $1346.1 \pm 16.4$ \\
 & \texttt{LinTS} & $2515.4 \pm 25.4$ & $2181.6 \pm 31.7$ & $1900.8 \pm 30.8$ & $1623.3 \pm 19.8$ & $1343.2 \pm 12.6$ \\
\cmidrule(lr){1-7}
\multirow{10}{*}{\shortstack[l]{Unknown\\Clusters}} & \underline{\texttt{(BTN)}\textsuperscript{2}} & \cellcolor{realdataSage}$\mathbf{2285.3 \pm 28.3}$ & \cellcolor{realdataSage}$\mathbf{1797.4 \pm 35.2}$ & \cellcolor{realdataSage}$\mathbf{1572.7 \pm 28.1}$ & \cellcolor{realdataSage}$\mathbf{1381.5 \pm 16.0}$ & \cellcolor{realdataSage}$\mathbf{1192.6 \pm 8.9}$ \\
 & \texttt{CLUB} & $2554.4 \pm 21.5$ & $2310.1 \pm 23.8$ & $2070.4 \pm 23.3$ & $1812.1 \pm 20.5$ & $1549.4 \pm 21.9$ \\
 & \texttt{SCLUB} & $2571.9 \pm 23.4$ & $2326.1 \pm 23.3$ & $2078.8 \pm 20.3$ & $1819.0 \pm 16.6$ & $1547.6 \pm 16.0$ \\
 & \texttt{CAB} & $2597.0 \pm 25.2$ & $2271.1 \pm 36.9$ & $1958.3 \pm 31.0$ & $1631.5 \pm 29.1$ & $1309.6 \pm 15.7$ \\
 & \texttt{COFIBA} & $2597.7 \pm 24.7$ & $2274.6 \pm 34.2$ & $1961.3 \pm 30.3$ & $1639.7 \pm 28.7$ & $1314.0 \pm 15.1$ \\
 & \texttt{RCLUMB} & $2853.3 \pm 22.6$ & $2770.9 \pm 26.1$ & $2742.8 \pm 24.8$ & $2724.3 \pm 28.4$ & $2700.3 \pm 30.2$ \\
 & \texttt{RSCLUMB} & $2851.1 \pm 22.5$ & $2770.6 \pm 25.7$ & $2745.1 \pm 22.3$ & $2723.9 \pm 23.5$ & $2710.5 \pm 28.5$ \\
 & \texttt{UniCLUB-E} & $2618.7 \pm 23.5$ & $2319.6 \pm 29.6$ & $2031.0 \pm 27.8$ & $1733.3 \pm 19.5$ & $1421.5 \pm 14.6$ \\
 & \texttt{UniCLUB-F} & $2815.6 \pm 17.3$ & $2821.1 \pm 17.1$ & $2819.4 \pm 14.2$ & $2817.3 \pm 15.6$ & $2816.3 \pm 16.2$ \\
 & \texttt{M-CNB} & $2478.6 \pm 26.9$ & $2082.0 \pm 28.1$ & $1769.4 \pm 31.8$ & $1489.6 \pm 21.0$ & $1218.1 \pm 17.1$ \\
\bottomrule
\end{tabular}
\endgroup
\end{table}

We additionally examine the clustering performance with unknown clusters in 
Table~\ref{tab:realdata-v4-two-ari}.
\texttt{(BTN)}\textsuperscript{2} achieves good cluster recovery, with mean ARI ranging from $0.784$ to $0.949$ and exact recovery in $72\%$--$92\%$ of runs.
Although some baselines achieve more accurate recovery, \texttt{(BTN)}\textsuperscript{2} has lower regret, which illustrates that more accurate clustering does not necessarily translate into better decision-making.

\begin{table}[htbp]
\centering
\caption{Cluster recovery after interaction.
We report mean final ARI, exact recovery counts, and mean global regret per user.
All baseline parameters are selected post hoc to maximize mean final ARI, whereas the choices minimizing global regret yield either a single cluster containing all users or one cluster per user.
\texttt{(BTN)}\textsuperscript{2} retains settings in \Cref{tab:realdata-v4-two-unknown}.
\textbf{Bold} marks the best result for each metric and $K$.}
\label{tab:realdata-v4-two-ari}
\begingroup\scriptsize\setlength{\tabcolsep}{2pt}
\resizebox{\textwidth}{!}{%
\begin{tabular}{lrrrrrrrrrrrrrrr}
\toprule
Method & \multicolumn{3}{c}{$K=2$} & \multicolumn{3}{c}{$K=4$} & \multicolumn{3}{c}{$K=6$} & \multicolumn{3}{c}{$K=8$} & \multicolumn{3}{c}{$K=10$} \\
\cmidrule(lr){2-4} \cmidrule(lr){5-7} \cmidrule(lr){8-10} \cmidrule(lr){11-13} \cmidrule(lr){14-16}
 & ARI & Exact & Global & ARI & Exact & Global & ARI & Exact & Global & ARI & Exact & Global & ARI & Exact & Global \\
\midrule
\underline{\texttt{(BTN)}\textsuperscript{2}} & 0.949 & 45/50 & \textbf{2285.3} & 0.928 & 44/50 & \textbf{1797.4} & 0.923 & \textbf{46/50} & \textbf{1572.7} & 0.876 & \textbf{43/50} & \textbf{1381.5} & 0.784 & \textbf{36/50} & \textbf{1192.6} \\
\texttt{CLUB} & \textbf{1.000} & \textbf{50/50} & 2577.7 & \textbf{1.000} & \textbf{50/50} & 2260.0 & \textbf{0.955} & 41/50 & 1957.3 & \textbf{0.937} & \textbf{43/50} & 1646.1 & 0.731 & 29/50 & 1324.4 \\
\texttt{SCLUB} & 0.665 & 10/50 & 2558.3 & 0.533 & 0/50 & 2247.0 & 0.464 & 0/50 & 1961.8 & 0.314 & 0/50 & 1669.8 & 0.244 & 0/50 & 1384.5 \\
\texttt{RSCLUMB} & 0.598 & 5/50 & 2560.2 & 0.341 & 0/50 & 2262.4 & 0.300 & 0/50 & 1973.5 & 0.277 & 0/50 & 1661.9 & 0.202 & 0/50 & 1404.4 \\
\texttt{UniCLUB-E} & 0.743 & 15/50 & 2577.4 & 0.520 & 26/50 & 2314.3 & 0.720 & 36/50 & 2029.4 & 0.820 & 41/50 & 1737.3 & \textbf{0.807} & 32/50 & 1436.0 \\
\texttt{UniCLUB-F} & 0.481 & 0/50 & 2815.6 & 0.322 & 14/50 & 2821.1 & 0.274 & 9/50 & 2819.4 & 0.292 & 7/50 & 2817.3 & 0.324 & 6/50 & 2816.3 \\
\bottomrule
\end{tabular}%
}
\endgroup
\end{table}

\section{Conclusion and Discussion}

We study online nonparametric stochastic bandits as a statistical framework for personalizing frozen black-box language models from user feedback.
We propose \texttt{(BTN)}\textsuperscript{2}, a nearest-neighbor UCB policy that achieves minimax-optimal regret rates.
We further introduce a distance between users' oracle choice distributions that captures shared preference rankings despite differences in their reward functions.
Based on this distance, we develop a strategy that detects user clusters and collaborates within clusters.
Experiments on a real question-answering benchmark demonstrate the effectiveness of our approach.

Several directions remain open. Our extension to dependent candidate sets quantifies how the joint density upper and lower bounds affect the regret bound.
Its assumptions remain restrictive, leaving room for further relaxation.
Another direction is to relax the main model's assumption that candidate sets are generated independently of the interaction history. In practice, contexts in a multi-turn conversation are rarely independent. Although martingale methods provide tools for analyzing history-dependent data, possible nonstationarity introduces additional challenges that require further modeling and methodological development.

\setlength{\bibsep}{1pt plus 1ex}
\bibliographystyle{apalike-pmlr}
\bibliography{ref}

\end{document}